\documentclass[11pt,twoside]{article}
\usepackage{geometry}
\usepackage{enumerate}
\usepackage{enumitem}
\usepackage{latexsym,booktabs}
\usepackage{amsmath,amssymb,bbold, xfrac}
\usepackage{graphicx}
\usepackage[singlespacing]{setspace}
\usepackage{algorithm}
\usepackage[noend]{algpseudocode}
\usepackage{multirow}

\usepackage{caption}

\usepackage{hyperref}
\hypersetup{colorlinks=True, urlcolor=cyan}

\usepackage[backend=bibtex, citestyle=authoryear, bibstyle=alphabetic]{biblatex}
\numberwithin{Theorem}{section}
\numberwithin{Definition}{section}
\numberwithin{Lemma}{section}
\numberwithin{Algorithm}{section}
\numberwithin{equation}{section}

\usepackage[frozencache, cachedir=./minted_files]{minted}
\usepackage{listings}
\usepackage{xcolor}
\newcommand{\pinline}[1]{\mintinline{python}{#1}}
\newminted{python}{framesep=2mm,
baselinestretch=1.2,
fontsize=\small,
frame=single
}

\newcommand{\V}{\mathbf{V}}
\newcommand{\Y}{\mathbf{Y}}
\newcommand{\hess}{\mathbf{H}}
\newcommand{\Thetab}{\mathbf{\Theta}}

\newcommand{\vb}{\mathbf{v}}
\newcommand{\yb}{\mathbf{y}}
\newcommand{\xb}{\mathbf{x}}

\newcommand{\jac}{\mathbf{J}}
\newcommand{\hessian}{\mathbf{H}}
\newcommand{\thetab}{\boldsymbol{\theta}}

\newcommand{\region}{B_{d,\epsilon}}
\newcommand{\indicator}[1]{\mathbb{1}_{#1}}
\newcommand{\regioni}{B_{d,\epsilon}^i}
\newcommand{\data}{\mathbf{y_0}}

\newcommand{\R}{\mathbb{R}}

\begin{document}

\pagestyle{empty}

% =============================================================================
% Title page
% =============================================================================
\begin{titlepage}
\vspace*{.5em}
\center
\textbf{\large{The School of Mathematics}} \\
\vspace*{1em}
\begin{figure}[!h]
\centering
\includegraphics[width=180pt]{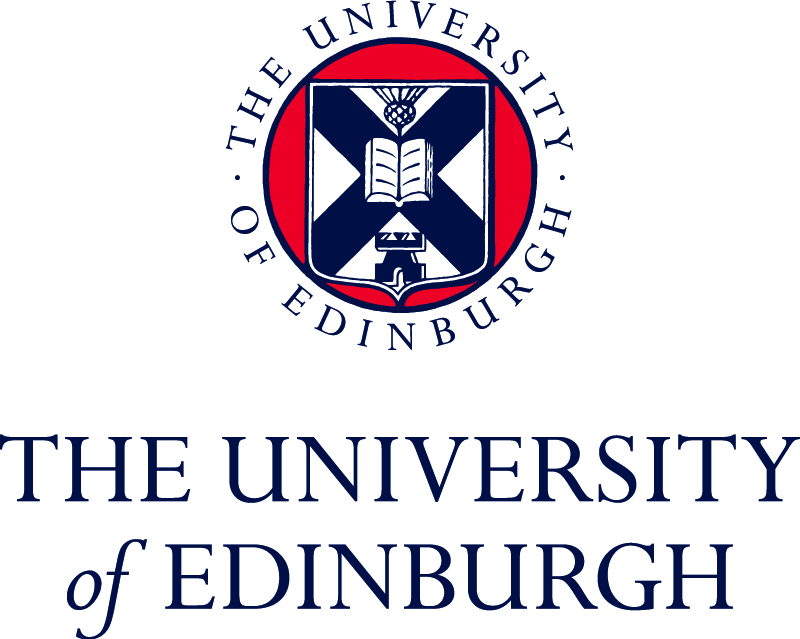}
\end{figure}
\vspace{2em}
\textbf{\Huge{Extending the statistical software package Engine for Likelihood-Free Inference}}\\[2em]
\textbf{\LARGE{by}}\\
\vspace{2em}
\textbf{\LARGE{Vasileios Gkolemis}}\\
\vspace{6.5em}
\Large{Dissertation Presented for the Degree of\\
MSc in Operational Research with Data Science}\\
\vspace{6.5em}
\Large{August 2020}\\
\vspace{3em}
\Large{Supervised by\\Dr. Michael Gutmann}
\vfill
\end{titlepage}

\cleardoublepage
% =============================================================================
% Abstract, acknowledgments, and own work declaration
% =============================================================================
\begin{center}
\Large{Abstract}
\end{center}

Bayesian inference is a principled framework for dealing with
uncertainty. The practitioner can
perform an initial assumption for the physical phenomenon they want to
model (prior belief), collect some data and then adjust the
initial assumption in the light of the new evidence (posterior
belief).  Approximate Bayesian Computation (ABC) methods, also known
as likelihood-free inference techniques, are a class of models used for
performing inference when the likelihood is intractable. The unique
requirement of these models is a black-box sampling machine. Due to
the modelling-freedom they provide these approaches are particularly
captivating.

Robust Optimisation Monte Carlo (ROMC) is one of the most recent
techniques of the specific domain. It approximates the posterior
distribution by solving independent optimisation problems. This
dissertation focuses on the implementation of the ROMC method in the
software package "Engine for Likelihood-Free Inference" (ELFI). In the
first chapters, we provide the mathematical formulation and the
algorithmic description of the ROMC approach. In the following
chapters, we describe our implementation; (a) we present all the
functionalities provided to the user and (b) we demonstrate how to
perform inference on some real examples.  Our implementation provides 
a robust and efficient solution to a
practitioner who wants to perform inference on a simulator-based
model. Furthermore, it exploits parallel processing for accelerating the inference wherever it is possible. Finally, it has been designed to serve extensibility; the
user can easily replace specific subparts of the method without
significant overhead on the development side. Therefore, it can be
used by a researcher for further experimentation.

\clearpage

\begin{center}
\Large{Acknowledgments}
\end{center}

I would like to thank Michael Gutmann, who was an excellent supervisor
throughout the whole period of the dissertation.  His directions and
insights were fundamental in the completion of this thesis. Despite
the difficulties of remote communication, our collaboration remained
pleasant and constructive.

Above all, a special thank to my family for supporting me all this
season. Without their support, I would not have made it to complete
this program.

\clearpage

\begin{center}
\Large{Own Work Declaration}
\end{center}

\hfill

I declare that this thesis was composed by myself and that
the work contained therein is my own, except where explicitly stated
otherwise in the text.

\hfill

\hfill

Vasileios Gkolemis
\cleardoublepage

% =============================================================================
% Table of contents, tables, and pictures (if applicable)
% =============================================================================
\pagestyle{plain}
\setcounter{page}{1}
\pagenumbering{Roman}

\tableofcontents
\clearpage
\listoftables
\listoffigures
\cleardoublepage

\pagenumbering{arabic}
\setcounter{page}{1}

\nocite{*}
\clearpage

%%%%%%%%%%%%%%%%%%%%%%%%%%%%%%%%%%%%%%%%
\section{Introduction}
\label{sec:introduction}
This dissertation is mainly focused on the implementation of the
Robust Optimisation Monte Carlo (ROMC) method as it was proposed by
\autocite{Ikonomov2019}, at the Python package ELFI - "Engine For
Likelihood-Free Inference" \autocite{1708.00707}. ROMC is a novel
likelihood-free inference approach for simulator-based models.

\subsection{Motivation}
\label{subsec:motivation}
\subsubsection*{\textit{Explanation of simulation-based models}}

A simulator-based model is a parameterised stochastic data generating
mechanism \autocite{Gutmann2016}. The key characteristic of these
models is that although we can sample (simulate) data points, we
cannot evaluate the likelihood of a specific set of observations
$\data$. Formally, a simulator-based model is described as a
parameterised family of probability density functions
$\{ p_{\yb|\thetab}(\yb) \}_{\thetab}$, whose closed-form is either
unknown or intractable to evaluate. Whereas evaluating
$p_{\yb|\thetab}(\yb)$ is intractable, sampling is
feasible. Practically, a simulator can be understood as a black-box
machine $M_r$\footnote{The subscript $r$ in $M_r$ indicates the
  \textit{random} simulator. In the next chapters we will introduce
  $M_d$ witch stands for the \textit{deterministic} simulator.} that
given a set of parameters $\thetab$, produces samples $\yb$ in a
stochastic manner i.e.\ $M_r(\thetab) \rightarrow \yb$.

Simulator-based models are particularly captivating due to the
modelling freedom they provide; any physical process that can be
conceptualised as a computer program of finite (deterministic or
stochastic) steps can be modelled as a simulator-based model without
any compromise. The modelling freedom includes any amount of hidden
(unobserved) internal variables or logic-based decisions. As always,
this degree of freedom comes at a cost; performing the inference is
particularly demanding from both computational and mathematical
perspective. Unfortunately, the algorithms deployed so far permit the
inference only at low-dimensionality parametric spaces, i.e.\
$\thetab \in \mathbb{R}^D$ where $D$ is small.

\subsubsection*{\textit{Example}}

\begin{figure}[!ht]
    \begin{center}
      \includegraphics[width=0.75\textwidth]{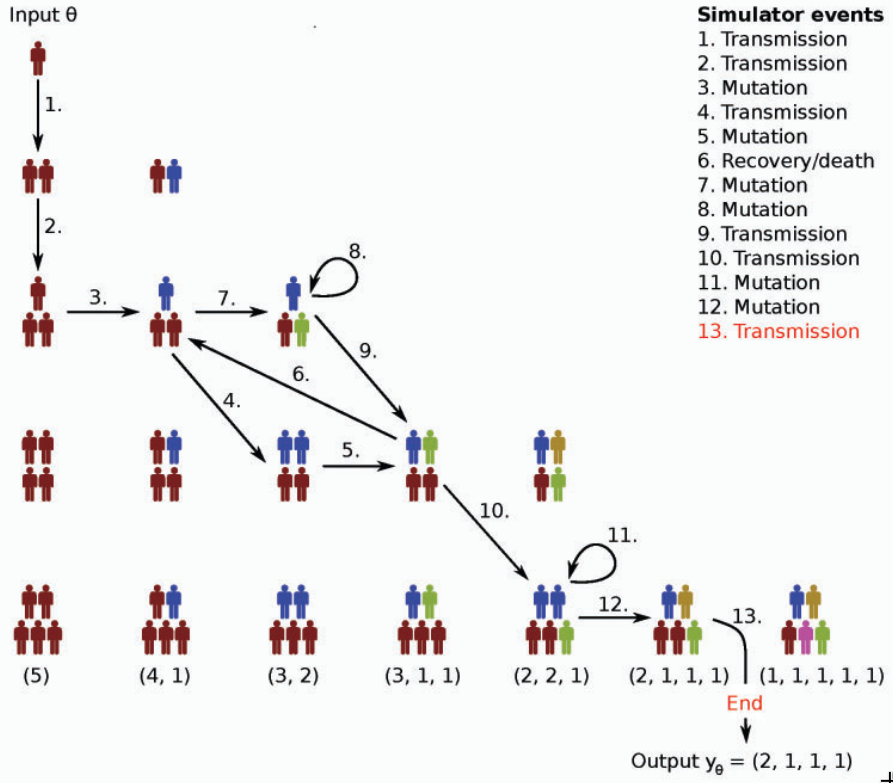}
    \end{center}
    \caption[The tuberculosis spreading process.]{Depiction of a random example from the tuberculosis
      spreading process. The image has been taken from
      \autocite{Lintusaari2017}.}
    \label{fig:tuberculosis_model}
\end{figure}

For underlying the importance of simulator-based models, let us use
the tuberculosis disease spread example as described in
\autocite{Tanaka2006}. An overview of the disease spread model is
presented at figure~\ref{fig:tuberculosis_model}. At each stage one of
the following \textit{unobserved} events may happen; (a) the
transmission of a specific haplotype to a new host, (b) the mutation
of an existent haplotype or (c) the exclusion of an infectious host
(recovers/dies) from the population. The random process, which stops
when $m$ infectious hosts are reached\footnote{We suppose that the
  unaffected population is infinite, so a new host can always be added
  until we reach $m$ simultaneous hosts.}, can be parameterised by the
transmission rate $\alpha$, the mutation rate $\tau$ and the exclusion
rate $\delta$, creating a $3D$-parametric space
$\thetab = (\alpha, \tau, \delta)$. The outcome of the process is a
variable-size tuple $\yb_{\thetab}$, containing the population
contaminated by each different haplotype, as described in
figure~\ref{fig:tuberculosis_model}. Let's say that the disease has
been spread in a real population and when $m$ hosts were contaminated
simultaneously, the vector with the infectious populations has been
measured to be $\data$. We would like to discover the parameters
$\thetab = (\alpha, \tau, \delta)$ that generated the spreading
process and led to the specific outcome $\data$. Computing
$p(\yb=\data|\thetab)$ requires tracking all tree-paths that could
generate the specific tuple; such exhaustive enumeration becomes
intractable when $m$ grows larger, as in real-case scenarios. In
figure~\ref{fig:tuberculosis_model} we can observe that a transmission
followed by a recovery/death creates a loop, reinstating the process
to the previous step, which also complicates the exhaustive
enumeration. Hence, representing the process with a simulator-based
model\footnote{which is simple and efficient} and performing
likelihood-free inference is the recommended solution.

\subsubsection*{\textit{Goal of Simulation-Based Models}}

As in most Machine Learning (ML) concepts, the fundamental goal is the
derivation of one(many) parameter configuration(s) $\thetab^*$ that
\textit{describe} the data best i.e.\ generate samples
$M_r(\thetab^*)$ that are as close as possible to the observed data
$\data$. In our case, following the approach of Bayesian ML, we treat
the parameters of interest $\thetab$ as random variables and we try to
\textit{infer} a posterior distribution $p(\thetab|\data)$ on them.

\subsubsection*{\textit{Robust Optimisation Monte Carlo (ROMC) method}}

The ROMC method~\autocite{Ikonomov2019} is very a recent
likelihood-free approach. Its fundamental idea is the transformation
of the stochastic data generation process $M_r(\thetab)$ to a
deterministic mapping $g_i(\thetab)$, by sampling the variables that
produce the randomness $\vb_i \sim p(\V)$. Formally, in every
stochastic process the randomness is influenced by a vector of random
variables $\V$, whose state is unknown prior to the execution of the
simulation; sampling the state makes the procedure deterministic,
namely $g_i(\thetab) = M_d(\thetab, \V=\vb_i)$. This approach
initially introduced at \autocite{Meeds2015} with the title
\textit{Optimisation Monte Carlo (OMC)}. The ROMC extended this
approach by resolving a fundamental failure-mode of OMC\@. The ROMC
describes a methodology for approximating the posterior through a
series of steps, without explicitly enforcing which algorithms must be
utilised for each step\footnote{The implementation chooses a specific
  algorithm for each task, but this choice has just a demonstrative
  value; any appropriate algorithm can be used instead.}; in this
sense, it can be perceived as a meta-algorithm.

\subsubsection*{\textit{Implementation}}

The most important contribution of this work is the implementation of
the ROMC method in the Python package Engine for Likelihood-Free
Inference (ELFI) \autocite{1708.00707}. Since the method has been
published quite recently, it has not been implemented by now in any ML
software. This work attempts to provide the research community with a
robust and extensible implementation for further experimentation.

\subsection{Outline of Thesis}
\label{subsec:outline-of-thesis}
The remainder of the dissertation is organised as follows; in Chapter
2, we establish the mathematical formulation. Namely, we initially
describe the simulator-based models and provide some background
information on the fundamental algorithms proposed so far. Afterwards,
we provide the mathematical description of the ROMC approach
\autocite{Ikonomov2019}. Finally, we transform the mathematical
description to algorithms. In Chapter 3, we illustrate the
implementation part; we initially provide some information regarding
the Python package Engine for Likelihood-Free Inference (ELFI)
\autocite{1708.00707} and subsequently, we present the implementation
details of ROMC in this package. In general, the conceptual scheme
followed by the dissertation is

$$\text{Mathematical modelling} \rightarrow
\text{Algorithm} \rightarrow \text{Software}$$

In Chapter 4, we demonstrate the functionalities of the ROMC
implementation at some real-world examples; this chapter demonstrates
the accuracy of the ROMC method and our implementation's at
likelihood-free tasks. Finally, in Chapter 5, we conclude with some
thoughts on the work we have done and some future research ideas.

\subsection{Notation}
\label{subsec:notation}
In this section, we provide an overview of the symbols utilised in the
rest of the document. At this level, the quantities are introduced
quite informally; most of them will be defined formally in the next
chapters. We try to keep the notation as consistent as possible
throughout the document. The symbol $\R^N$, when used, describes that
a variable belongs to the $\text{N-dimensional}$ Euclidean space; $N$
does not represent a specific number. The bold formation ($\xb$)
indicates a vector, while ($x$) a scalar. Random variables are
represented with capital letters ($\Thetab$) while the samples with
lowercase letters ($\thetab$), i.e.\ $\thetab \sim \Thetab$.

\subsubsection*{Random Generator}
\label{sec:random-generator}
\begin{itemize}
\item $M_r(\thetab): \R^D \rightarrow \R$: The black-box data simulator
\end{itemize}

\subsubsection*{Parameters/Random Variables/Symbols}
\label{sec:variables}

\begin{itemize}
\item $D \in \mathbb{N}$, the dimensionality of the parameter-space
\item $\Thetab \in \R^D$, random variable representing the parameters of interest
\item $\data \in \R^N$, the vector with the observations
\item $\epsilon \in R$, the threshold setting the limit on the region
  around $\data$. When further notation is introduced
  regarding $\epsilon$\footnote{If
    $\epsilon_*, \: *:\{filter, region, cutoff\}$ values are not specified
    explicitly, they all share the common value of $\epsilon$}
  \begin{itemize}
  \item $\epsilon_{filter}$, threshold for discarding solutions
  \item $\epsilon_{region}$, threshold for building bounding box regions
  \item $\epsilon_{cutoff}$, threshold for the indicator function
  \end{itemize}
  
\item $\V \in \R^N$, random variable representing the randomness of
  the generator. It is also called nuisance variable, because we are
  not interested in inferring a posterior distribution on it.
\item $\vb_i \sim \V$, a specific sample drawn from $\V$
\item $\Y_{\thetab}$, random variable describing the simulator $M_r(\thetab)$. 
\item $\yb_i \sim \Y_\theta$, a sample drawn from $\Y_\theta$. It can
  be obtained by executing the simulator $\yb_i \sim M_r(\thetab)$
\end{itemize}

\subsubsection*{Sets}
\label{sec:sets}

\begin{itemize}
\item $\region(\data)$, the set of $\yb$ points close to the
  observations, i.e.\ $\yb := \{\yb: d(\yb, \data) \leq \epsilon \}$
\item $\regioni$, the set of points defined around $\yb_i$ i.e.\
  $\regioni = \region(\yb_i)$
\item $\mathcal{S}_i$, the set of parameters $\thetab$ that generate data close
  to the observations using the i-th deterministic generator, i.e.\
  $\{ \thetab: M_d(\thetab, \vb_i) \in \region(\data) \}$
\end{itemize}
    
\subsubsection*{Generic Functions}
\label{sec:generic-functions}

\begin{itemize}
\item $p(\cdot)$, any valid pdf
\item $p(\cdot | \cdot)$, any valid conditional distribution
\item $p(\thetab)$, the prior distribution on the parameters
\item $p(\vb)$, the prior distribution on the nuisance variables
\item $p(\thetab|\data)$, the posterior distribution
\item $p_{d,\epsilon}(\thetab|\data)$, the approximate posterior
  distribution  
\item $d(\mathbf{x}, \mathbf{y}): \R^{2N} \rightarrow \R$: any valid
  distance, the $L_2$ norm: $||\mathbf{x}-\mathbf{y}||_2$
\end{itemize}

\subsubsection*{Functions (Mappings)}
\label{sec:functions-mappings}

\begin{itemize}
\item $M_d(\thetab, \vb): \R^D \rightarrow \R$, the deterministic
  generator; all stochastic variables that are part of the data generation process are represented by the parameter $\vb$
\item $f_i(\thetab) = M_d(\thetab, \vb_i)$, deterministic generator associated with sample $\vb_i \sim p(\vb)$
\item $g_i(\thetab) = d(f_i(\thetab), \data)$, distance of the generated data $f_i(\thetab)$ from the observations
\item $T(\mathbf{x}): \mathbb{R}^{D_1} \rightarrow \mathbb{R}^{D_2}$
  where $D_1 > D_2$, the mapping that computes the summary statistic
\item $\indicator{\region(\data)}(\yb)$, the indicator function; returns 1 if $d(\yb, \data) \leq \epsilon$, else 0
\item $L(\thetab)$, the likelihood
\item $L_{d,\epsilon}(\thetab)$, the approximate likelihood
\end{itemize}

\subsection{Online notebooks}
\label{subsec:notebooks}
Since the dissertation is mainly focused on the implementation of the
ROMC, we have created some interactive notebooks\footnote{We have used
  the jupyter notebooks \autocite{Kluyver:2016aa}} supporting the main
document. The reader is advised to exploit the notebooks in order to
(a) review the code used for performing the inference (b) understand
the functionalities of our implementation in a practical sense (c)
check the validity of our claims and (c) interactively execute
experiments.

The notebooks are provided in two formats:\footnote{The
  notebooks in the two repositories are identical.}

\begin{itemize}
\item The first format is located in
  \href{https://github.com/givasile/edinburgh-thesis/tree/master/notebook_examples}{this}
  Github repository. In order for the reader to experiment with the
  examples, they should clone the repository into their pc, execute
  the installation instructions and run the notebooks using the
  Jupyter package.
\item The second format is the google colab notebook
  \autocite{Bisong2019}. Following the links below, the reader can
  view the notebook and make an online clone of it. Hence, without any
  installation overhead, they may interactivelly explore all the
  provided functionalities.
\end{itemize}

The notebooks provide a practical overview of the implemented
functionalities and it are easy to use; in particular, the google
colab version is entirely plug-and-play. Therefore, we encourage the
reader of the dissertation to use them as supporting material.  The
following list contains the implemented examples along with their
links:

\begin{itemize}
\item Simple 1D example
  \begin{itemize}
  \item \href{https://colab.research.google.com/drive/1lGRp0XrNfZ64NN0ASB_tYEKowXwlveDC?usp=sharing}{Google colab}
  \item \href{https://github.com/givasile/edinburgh-thesis/blob/master/notebook_examples/example_1D.ipynb}{Github repository}
  \end{itemize}
 
\item Simple 2D example
  \begin{itemize}
  \item
    \href{https://colab.research.google.com/drive/1T8919FCAi2w9MXm9XKT_iJLnB0y1EN32?usp=sharing}{Google colab}
  \item \href{https://github.com/givasile/edinburgh-thesis/blob/master/notebook_examples/example_2D.ipynb}{Github repository}
  \end{itemize}

\item Moving Average example  
  \begin{itemize}
  \item
    \href{https://colab.research.google.com/drive/1nkdACQ370SSc0KB1bHv4sBRaxMlMqoNH?usp=sharing}{Google colab}
  \item
    \href{https://github.com/givasile/edinburgh-thesis/blob/master/notebook_examples/example_MA2.ipynb}{Github repository}
  \end{itemize}

\item Extensibility example
  \begin{itemize}
  \item
    \href{https://colab.research.google.com/drive/1RzB-V1QueP1y1nyzv_VOqR1nVz3DUH3v?usp=sharing}{Google colab}
  \item
    \href{https://github.com/givasile/edinburgh-thesis/blob/master/notebook_examples/extending_romc.ipynb}{Github repository}
  \end{itemize}

\end{itemize}

\clearpage

%%%%%%%%%%%%%%%%%%%%%%%%%%%%%%%%%%%%%%%%
\section{Background}
\label{sec:background}

\subsection{Simulator-based models}
As already stated at Chapter~\ref{sec:introduction}, in
simulator-based models we cannot evaluate the posterior
$p(\thetab|\data) \propto L(\thetab)p(\thetab)$, due to the
intractability of the likelihood $L(\thetab) = p(\data|\thetab)$. The
following equation allows incorporating the simulator in the place of
the likelihood and forms the basis of all likelihood-free inference
approaches,

\begin{equation} \label{eq:likelihood}
  L(\thetab) =
  \lim_{\epsilon \to 0} c_\epsilon \int_{\yb \in B_{d,\epsilon}(\data)} p(\yb|\thetab)d\yb =
  \lim_{\epsilon \to 0} c_\epsilon \Pr(M_r(\thetab) \in B_{d,\epsilon}(\data))
\end{equation}
where $c_\epsilon$ is a proportionality factor dependent on
$\epsilon$, needed when
$\Pr(M_r(\thetab) \in B_{d, \epsilon}(\data)) \rightarrow 0$, as
$\epsilon \rightarrow 0$. Equation~\ref{eq:likelihood} describes that
the likelihood of a specific parameter configuration $\thetab$ is
proportional to the probability that the simulator will produce
outputs equal to the observations, using this configuration.

\subsubsection{Approximate Bayesian Computation (ABC) Rejection
  Sampling}

ABC rejection sampling is a modified version of the traditional
rejection sampling method, for cases when the evaluation of the
likelihood is intractable. In the typical rejection sampling, a sample
obtained from the prior $\thetab \sim p(\thetab)$ gets accepted
with probability $L(\thetab)/ \text{max}_{\thetab}
L(\thetab)$. Though we cannot use this approach out-of-the-box
(evaluating $L(\thetab)$ is impossible in our case), we can
modify the method incorporating the simulator.

In the discrete case scenario where $\Y_{\thetab}$ can take a finite
set of values, the likelihood becomes
$L(\thetab) = \Pr(\Y_{\thetab} = \data)$ and the posterior
$p(\thetab|\data) \propto \Pr(\Y_{\thetab}=\data)p(\thetab)$; hence, we can
sample from the prior $\thetab_i \sim p(\thetab)$, run the simulator
$\yb_i = M_r(\thetab_i)$ and accept $\thetab_i$ only if
$\yb_i = \data$.

The method above becomes less useful as the finite set of
$\Y_{\thetab}$ values grows larger, since the probability of
accepting a sample becomes smaller. In the limit
where the set becomes infinite (i.e.\ continuous case) the probability
becomes zero. In order for the method to work in this set-up, a
relaxation is introduced; we relax the acceptance criterion by letting
$\yb_{i}$ lie in a larger set of points i.e.\
$\yb_{i} \in \region(\data), \epsilon > 0$. The region can be
defined as $\region (\data) := \{\yb: d(\yb, \data) \leq \epsilon \}$
where $d(\cdot, \cdot)$ can represent any valid distance. With this
modification, the maintained samples follow the approximate posterior,

\begin{equation} \label{eq:approx_posterior}
  p_{d,\epsilon}(\thetab|\data) \propto Pr(\yb \in
  \region(\data)) p(\thetab)
\end{equation}

\noindent
This method is called Rejection ABC.

\subsubsection{Summary Statistics}

When $\yb \in \mathbb{R}^D$ lies in a high-dimensional space, generating
samples inside $\region (\data)$ becomes rare even when $\epsilon$ is
relatively large; this is the curse of dimensionality. As a
representative example lets make the following hypothesis;

\begin{itemize}
\item $d$ is set to be the Euclidean distance, hence
  $\region(\data) := \{ \yb: ||\yb - \data||_2^2 < \epsilon^2 \}$ is a
  hyper-sphere with radius $\epsilon$ and volume $V_{hypersphere} = \frac{\pi^{D/2}}{\Gamma(D/2 + 1)} \epsilon^D$
\item the prior $p(\thetab)$ is a uniform distribution in a hyper-cube with side of
  length $2\epsilon$ and volume $V_{hypercube} = (2\epsilon)^D$
\item the generative model is the identity function $\yb=f(\thetab)= \thetab $
\end{itemize}

\noindent
The probability of drawing a sample inside the hypersphere equals the
fraction of the volume of the hypersphere inscribed in the hypercube:

\begin{equation}
  Pr(\yb \in \region (\data))
  = Pr(\thetab \in \region (\data))
  = \frac{V_{hypersphere}}{V_{hypercube}}
  = \frac{\pi^{D/2}}{2^D\Gamma(D/2 + 1)} \rightarrow 0, \quad \text{as} \quad D \rightarrow \infty
\end{equation}

\noindent
We observe that the probability tends to $0$, independently of
$\epsilon$; enlarging $\epsilon$ will not increase the acceptance
rate. Intuitively, we can think that in high-dimensional spaces the
volume of the hypercube concentrates at its corners. This generates
the need for a mapping
$T: \mathbb{R}^{D_1} \rightarrow \mathbb{R}^{D_2}$ where $D_1 > D_2$,
for squeezing the dimensionality of the output. This
dimensionality-reduction step that redefines the area as
$\region(\data) := \{\yb: d(T(\yb), T(\data)) \leq \epsilon \}$ is
called \textit{summary statistic} extraction, since the distance is
not measured on the actual outputs, but on a summarisation (i.e.\
lower-dimension representation) of them.

\subsubsection{Approximations introduced so far}

So far, we have introduced some approximations for inferring the
posterior as
$p_{d,\epsilon}(\thetab|\data) \propto Pr(\Y_{\thetab} \in
\region(\data))p(\thetab)$ where
$\region(\data) := \{\yb: d(T(\yb), T(\data)) < \epsilon \}$. These
approximations introduce two different types of errors:

\begin{itemize}
\item $\epsilon$ is chosen to be \textit{big enough}, so that enough
  samples are accepted. This modification leads to the approximate
  posterior introduced in \eqref{eq:approx_posterior}
\item $T$ introduces some loss of information, making possible a $\yb$
  far away from $\data$ i.e.\ $\yb: d(\yb,\data)>\epsilon$, to enter
  the acceptance region after the dimensionality reduction
  $d(T(\yb), T(\data)) \leq \epsilon$
\end{itemize}

\noindent
In the following sections, we will not use the summary statistics in
our expressions for the notation not to clutter. One could understand
it as absorbing the mapping $T(\cdot)$ inside the simulator. In any
case, all the propositions that will be expressed in the following
sections are valid with the use of summary statistics.
  
\subsubsection{Optimisation Monte Carlo (OMC)}

Before we define the likelihood approximation as introduced in the
OMC, approach lets define the indicator function based on
$\region(\yb)$. The indicator function $\indicator{\region(\yb)}(\xb)$
returns 1 if $\xb \in \region(\yb)$ and 0 otherwise. If
$d(\cdot,\cdot)$ is a formal distance, due to symmetry
$\indicator{\region(\yb)}(\xb) = \indicator{\region(\xb)}(\yb)$, so
the expressions can be used interchangeably.

\begin{gather} \label{eq:indicator} \indicator{\region(\yb)}(\xb)=
  \left\{
    \begin{array}{ll}
      1 & \mbox{if } \xb \in \region(\yb) \\
      0 & \mbox{otherwise} 
    \end{array} \right. \end{gather}

\noindent
Based on equation~\eqref{eq:approx_posterior} and the indicator
function as defined above~\eqref{eq:indicator}, we can approximate the
likelihood as:

\begin{gather} \label{eq:approx_likelihood}
  L_{d, \epsilon}(\thetab) =
  \int_{\yb \in B_\epsilon(\data)}p(\yb|\thetab)d\yb =
  \int_{\yb \in \R^D} \indicator{\region(\data)}(\yb)p(\yb|\thetab)d\yb\\
  \approx \frac{1}{N} \sum_i^N \indicator{\region(\data)}(\yb_i),\text{ where }
  \yb_i \sim M_r(\thetab) \label{eq:init_view}\\
  \approx \frac{1}{N} \sum_i^N \indicator{\region (\data)} (\yb_i)
  \text{ where } \yb_i = M_d(\thetab, \vb_i), \vb_i \sim p(\vb) \label{eq:alt_view}
\end{gather}
This approach is quite intuitive; approximating the likelihood of a
specific $\thetab$ requires sampling from the data generator and count
the fraction of samples that lie inside the area around the
observations. Nevertheless, by using the approximation of equation
\eqref{eq:init_view} we need to draw $N$ new samples for each distinct
evaluation of $L_{d,\epsilon}(\thetab)$; this makes this approach
quite inconvenient from a computational point-of-view. For this
reason, we choose to approximate the integral as in equation
\eqref{eq:alt_view}; the nuisance variables are sampled once
$\vb_i \sim p(\vb)$ and we count the fraction of samples that lie
inside the area using the deterministic simulators
$M_d(\thetab, \vb_i) \: \forall i$. Hence, the evaluation
$L_{d,\epsilon}(\thetab)$ for each different $\thetab$ does not imply
drawing new samples all over again. Based on this approach, the
unnormalised approximate posterior can be defined as:

\begin{equation} \label{eq:aprox_posterior}
  p_{d,\epsilon}(\thetab|\data)
  \propto p(\thetab) \sum_i^N \indicator{ \region(\data)} (\yb_i)
\end{equation}

\subsubsection*{Further approximations for sampling and computing expectations}

The posterior approximation in \eqref{eq:approx_posterior} does not
provide any obvious way for drawing samples. In fact, the set
$\mathcal{S}_i = \{ \thetab: M_d(\thetab, \vb_i) \in \region(\data) \}$ can
represent any arbitrary shape in the D-dimensional Euclidean space; it
can be non-convex, can contain disjoint sets of $\thetab$ etc. We need
some further simplification of the posterior for being able to draw
samples from it.

As a side-note, weighted sampling could be performed in a
straightforward fashion with importance sampling. Using the prior as
the proposal distribution $\thetab_i \sim p(\thetab)$ and we can
compute the weight as
$w_i = \frac{L_{d,\epsilon}(\thetab_i)}{p(\thetab_i)}$, where
$L_{d,\epsilon}(\thetab_i)$ is computed with the expression
\eqref{eq:approx_likelihood}. This approach has the same drawbacks as
ABC rejection sampling; when the prior is wide or the dimensionality
$D$ is high, drawing a sample with non-zero weight is rare, leading to
either poor Effective Sample Size (ESS) or huge execution time.

The OMC proposes a quite drastic simplification of the posterior; it
squeezes all regions $\mathcal{S}_i$ into a single point
$\thetab_i^* \in \mathcal{S}_i$ attaching a weight $w_i$ proportional
to the volume of $\mathcal{S}_i$. For obtaining a
$\thetab_i^* \in \mathcal{S}_i$, a gradient based optimiser is used
for minimising $g_i(\thetab) = d(\data, f_i(\thetab))$ and the
estimation of the volume of $\mathcal{S}_i$ is done using the Hessian
approximation $\hess_i \approx \jac_i^{*T}\jac_i^*$, where $\jac_i^*$
is the Jacobian matrix of $g_i(\thetab)$ at $\thetab_i^*$. Hence,

\begin{gather} \label{eq:OMC_posterior}
    p(\thetab|\data) \propto p(\thetab) \sum_i^N w_i \delta(\thetab - \thetab_i^*)\\
  \thetab_i^* = \text{argmin}_{\thetab} \:g_i(\thetab) \\
  w_i \propto \frac{1}{\sqrt{det( \jac_i^{*T}\jac_i^*)}}
\end{gather}

The distribution \eqref{eq:OMC_posterior} provides weighted samples
automatically and an expectation can be computed easily with the
following equation,

\begin{equation}
  \label{eq:OMC_expectation}
  E_{p(\thetab|\data)}[h(\thetab)] = \frac{\sum_i^N w_i p(\thetab_i^*)h(\thetab_i^*)}{\sum_i^N w_i p(\thetab_i^*)}
\end{equation}

\subsection{Robust Optimisation Monte Carlo (ROMC) approach}
\label{subsec:ROMC}
The simplifications introduced by OMC, although quite useful from a
computational point-of-view, they suffer from some significant failure modes:

\begin{itemize}
\item The whole acceptable region $\mathcal{S}_i$, for each nuisance variable,
  shrinks to a single point $\thetab_i^*$; this simplification may add
  significant error when then the area $\mathcal{S}_i$ is relatively big.
\item The weight $w_i$ is computed based only at the curvature at the
  point $\thetab_i^*$. This approach is error prone at many cases
  e.g.\ when $g_i$ is almost flat at $\thetab_i^*$, leading to a
  $\text{det}(\jac_i^{*T}\jac_i^*) \rightarrow 0 \Rightarrow w_i
  \rightarrow \infty$, thus dominating the posterior.
\item There is no way to solve the optimisation problem
  $\thetab_i^* = \text{argmin}_{\thetab} \: g_i(\thetab)$ when $g_i$
  is not differentiable.
\end{itemize}

\subsubsection{Sampling and computing expectation in ROMC}

The ROMC approach resolves the aforementioned issues. Instead of
collapsing the acceptance regions $\mathcal{S}_i$ into single points,
it tries to approximate them with a bounding box.\footnote{The
  description on how to estimate the bounding box is provided in the
  following chapters.}. A uniform distribution is then defined on the
bounding box area, used as the proposal distribution for importance
sampling. If we define as $q_i$, the uniform distribution defined on
the $i-th$ bounding box, weighted sampling is performed as:

\begin{gather}
  \label{eq:sampling}
  \thetab_{ij} \sim q_i \\
  w_{ij} = \frac{\indicator{\region(\data)}(M_d(\thetab_{ij}, \vb_i)) p(\thetab_{ij})}{q(\thetab_{ij})}
\end{gather}

\noindent
Having defined the procedure for obtaining weighted samples, any
expectation $E_{p(\thetab|\data)}[h(\thetab)]$, can be approximated
as,

\begin{equation} \label{eq:expectation}
  E_{p(\thetab|\data)}[h(\thetab)] \approx \frac{\sum_{ij} w_{ij} h(\thetab_{ij})}{\sum_{ij} w_{ij}}
\end{equation}

\subsubsection{Construction of the proposal region}

In this section we will describe mathematically the steps needed for
computing the proposal distributions $q_i$. There will be also
presented a Bayesian optimisation alternative when gradients are not
available.

\subsubsection*{Define and solve deterministic optimisation problems}

For each set of nuisance variables $\vb_i, i = \{1,2,\ldots,n_1 \}$ a
deterministic function is defined as
$f_i(\thetab) = M_d(\thetab,\vb_i)$. For constructing the proposal
region, we search for a point
$\thetab_* : d(f_i(\thetab_*), \data) < \epsilon$; this point can be
obtained by solving the the following optimisation problem:

\begin{subequations}
\begin{alignat}{2}
&\!\min_{\thetab}        &\qquad& g_i(\thetab) = d(\data,  f_i(\thetab))\label{eq:optProb}\\
&\text{subject to} &      & g_i(\thetab) \leq \epsilon
\end{alignat}
\end{subequations}
We maintain a list of the solutions $\thetab_i^*$ of the optimisation
problems. If for a specific set of nuisance variables $\vb_i$, there
is no feasible solution we add nothing to the list. The optimisation
problem can be treated as unconstrained, accepting the optimal point
$\thetab_i^* = \text{argmin}_{\thetab} g_i(\thetab)$ only if
$g_i(\thetab_i^*) < \epsilon$.

\subsubsection*{Gradient-based approach}
\label{subsubsec:GB_approach}

The nature of the generative model $M_r(\thetab)$, specifies the
properties of the objective function $g_i$. If $g_i$ is continuous
with smooth gradients $\nabla_{\thetab} g_i$ any gradient-based
iterative algorithm can be used for solving~\ref{eq:optProb}. The
gradients $\nabla_{\thetab} g_i$ can be either provided in closed form
or approximated by finite differences.

\subsubsection*{Bayesian optimisation approach}
\label{subsubsec:GP_approach}

In cases where the gradients are not available, the Bayesian
optimisation scheme provides an alternative
choice~\autocite{Shahriari2016}. With this approach, apart from
obtaining an optimal $\thetab_i^* $, a surrogate model $\hat{d}_i$ of
the distance $g_i$ is fitted; this approximate model can be used in
the following steps for making the method more
efficient. Specifically, in the construction of the proposal region
and in
equations~\eqref{eq:approx_posterior},~\eqref{eq:sampling},~\eqref{eq:expectation}
it could replace $g_i$ in the evaluation of the indicator
function, providing a major speed-up.

\subsubsection*{Construction of the proposal area $q_i$}

After obtaining a $\thetab_i^*$ such that
$g_i(\thetab_i^*) < \epsilon$, we need to construct a bounding box
around it. The bounding box $\mathcal{\hat{S}}_i$ must contain the
acceptance region around $\thetab_i^*$, i.e.\
$\{ \thetab : g_i(\thetab) < \epsilon$,
$d(\thetab, \thetab_i^*) < M \}$. The second condition
$d(\thetab, \thetab_i^*) < M$ is meant to describe that if
$\mathcal{S}_i := \{ \thetab : g_i(\thetab) < \epsilon \} $ contains a
number of disjoint sets of $\thetab$ that respect
$g_i(\thetab) < \epsilon$, we want our bounding box to fit only the
one that contains $\thetab_i^*$. We seek for a bounding box that is as
tight as possible to the local acceptance region (enlarging the
bounding box without a reason decreases the acceptance rate) but large
enough for not discarding accepted areas.

In contrast with the OMC approach, we construct the bounding box by
obtaining search directions and querying the indicator function as we move on them. The
search directions $\mathbf{v}_d$ are computed as the eigenvectors of
the curvature at $\thetab_i^*$ and a line-search method is used to
obtain the limit point where
$g_i(\thetab_i^* + \kappa \vb_d) \geq
\epsilon$\footnote{$-\kappa$ is used as well for the opposite direction along the search line}. The Algorithm~\ref{alg:region_construction} describes the method in-depth. After the limits are obtained along all
search directions, we define bounding box and the uniform distribution $q_i$. This is the proposal distribution used for the importance
sampling as explained in \eqref{eq:sampling}.

\subsubsection*{Fitting a local surrogate model $\hat{g}_i$}

After the construction of the bounding box $\mathcal{\hat{S}}_i$, we
are no longer interested in the surface outside the box. In the future
steps (e.g. sampling, evaluating the posterior) we will only evaluate
$g_i$ inside the corresponding bounding box. Hence, we could fit a
local surrogate model $\hat{g}_i$ for representing the local area
around $\theta_i^*$. Doing so, in the future steps we can exploit
$\hat{g}_i$ for evaluating the indicator function instead of running
the whole deterministic simulator.

Any ML regression model may be chosen as local surrogates. The choice
should consider the properties of the local region (i.e. size,
smoothness). The ROMC proposes fitting a simple quadratic model. The
training set $X: \mathbb{R}^{N \times D}$ is created by sampling $N$
points from the bounding box and the labels $Y: \mathbb{R}^{N}$ are
the computed by evaluating $g_i$. The quadratic model is fitted on the
data points, for minimising the square error.

This additional step places an
additional step in the training part, increasing the computational
demands, but promises a major speed at the inference phase (sampling,
posterior evaluation). It is frequent in ML problems, to be quite
generous with the execution time at the training phase, but quite
eager at the inference phase. Fitting a local surrogate model aligns
with this requirement.

\subsection{Algorithmic description of ROMC}
\label{subsec:romc-algorithmic}
In this section, we will provide the algorithmic description of the
ROMC method; how to solve the optimisation problems using either the
gradient-based approach or the Bayesian optimisation alternative and
the construction of the bounding box. Afterwards, we will discuss the
advantages and disadvantages of each choice in terms of accuracy and
efficiency.

At a high-level, the ROMC method can be split into the training and
the inference part.

\subsubsection*{Training part}
\noindent
At the training (fitting) part, the goal is the estimation of the
proposal regions $q_i$. The tasks are (a) sampling the nuisance
variables $\vb_i \sim p(\vb)$ (b) defining the optimisation problems
$\min_{\thetab} \: g_i(\thetab)$ (c) obtaining $\thetab_i^*$ (d)
checking whether $d_i^* \leq \epsilon$ and (e) building the bounding
box for obtaining the proposal region $q_i$. If gradients are
available, using a gradient-based method is advised for obtaining
$\thetab_i^*$ much faster. Providing $\nabla_{\thetab} g_i$ in
closed-form provides an upgrade in both accuracy and efficiency; If
closed-form description is not available, approximate gradients with
finite-differences requires two evaluations of $g_i$ for
\textbf{every} parameter $\thetab_d$, which works adequately well for
low-dimensional problems. When gradients are not available or $g_i$ is
not differentiable, the Bayesian optimisation paradigm exists as an
alternative solution. In this scenario, the training part becomes
slower due to fitting of the surrogate model and the blind
optimisation steps. Nevertheless, the subsequent task of computing the
proposal region $q_i$ becomes faster since $\hat{d}_i$ can be used
instead of $g_i$; hence we avoid to run the simulator
$M_d(\thetab, \vb_i)$ for each query. The
algorithms~\ref{alg:training_GB} and~\ref{alg:training_GP} present the
above procedure.

\subsubsection*{Inference Part}
Performing the inference includes one or more of the following three
tasks; (a) evaluating the unnormalised posterior
$p_{d, \epsilon}(\thetab|\data)$ (b) sampling from the posterior
$ \thetab_i \sim p_{d, \epsilon}(\thetab|\data)$ (c) computing an
expectation $E_{\thetab|\data}[h(\thetab)]$. Computing an expectation
can be done easily after weighted samples are obtained using the
equation~\ref{eq:expectation}, so we will not discuss it separately.

\noindent
Evaluating the unnormalised posterior requires solely the
deterministic functions $g_i$ and the prior distribution $p(\thetab)$;
there is no need for solving the optimisation problems and building
the proposal regions. The evaluation requires iterating over all $g_i$
and evaluating the distance from the observed data. In contrast, using
the GP approach, the optimisation part should be performed first for
fitting the surrogate models $\hat{d}_i(\thetab)$ and evaluate the
indicator function on them. This provides an important speed-up,
especially when running the simulator is computationally
expensive. % The evaluation of the posterior is presented analytically
% in~\ref{alg:posterior_GB} and~\ref{alg:posterior_GP}.

\noindent
Sampling is performed by getting $n_2$ samples from each proposal
distribution $q_i$. For each sample $\thetab_{ij}$, the indicator
function is evaluated $\indicator{\regioni(\data)}(\thetab_{ij})$ for
checking if it lies inside the acceptance region. If so the
corresponding weight is computed as in \eqref{eq:sampling}. As before,
if a surrogate model $\hat{d}$ is available, it can be utilised for
evaluating the indicator function. At the sampling task, the
computational benefit of using the surrogate model is more valuable
compared to the evaluation of the posterior, because the indicator
function must be evaluated for a total of $n_1 \times n_2$ points.

\noindent
In summary, we can state that the choice of using a Bayesian
optimisation approach provides a significant speed-up in the inference
part with the cost of making the training part slower and a possible
approximation error. It is typical in many Machine-Learning use cases,
being able to provide enough time and computational resources for the
training phase, but asking for efficiency in the inference
part.

\begin{minipage}{0.46\textwidth}
\begin{algorithm}[H]
    \centering
    \caption{Training Part - Gradient-based. Requires $g_i(\thetab), p(\thetab)$}\label{alg:training_GB}
    \begin{algorithmic}[1]
      \For{$i \gets 1 \textrm{ to } n$}
        \State Obtain $\thetab_i^*$ using a Gradient Optimiser
        \If{$g_i(\thetab_i^*) > \epsilon$}
        \State{go to} 1
        \Else
        \State Approximate $\jac_i^* = \nabla g_i(\theta)$ and $H_i \approx \jac^T_i\jac_i$
        \State Use Algorithm~\ref{alg:region_construction} to obtain $q_i$
        \EndIf      
      \EndFor
      \Return{$q_i, p(\theta), g_i(\theta)$}
    \end{algorithmic}
\end{algorithm}
\end{minipage}
\hfill
\begin{minipage}{0.46\textwidth}
\begin{algorithm}[H]
    \centering
    \caption{Training Part - Bayesian optimisation. Requires $g_i(\thetab), p(\thetab)$}\label{alg:training_GP}
    \begin{algorithmic}[1]
      \For{$i \gets 1 \textrm{ to } n$}
        \State Obtain $\thetab_i^*, \hat{d}_i(\thetab)$ using a GP approach
        \If{$g_i(\thetab_i^*) > \epsilon$}
        \State{go to} 1
        \Else
        \State Approximate $H_i \approx \jac^T_i \jac_i$
        \State Use Algorithm~\ref{alg:region_construction} to obtain $q_i$
        \EndIf      
      \EndFor
      \Return{$q_i, p(\theta), \hat{d}_i(\theta)$}
    \end{algorithmic}
\end{algorithm}
\end{minipage}

\begin{algorithm}[!ht]
	\caption{Computation of the proposal distribution $q_i$; Needs, a model of distance $d$, optimal point $\thetab_i^*$, number of refinements $K$, step size $\eta$ and curvature matrix $\hessian_i$ ($\jac_i^T\jac_i $ or GP Hessian)}\label{alg:region_construction}
	\begin{algorithmic}[1]
	\State Compute eigenvectors $\vb_{d}$ of $\hess_i$ {\scriptsize ($d = 1,\ldots,||\thetab ||)$}
	\For{$d \gets 1 \textrm{ to } ||\thetab||$}
		\State $\Tilde{\thetab} \gets \thetab_i^*$ \label{algstep:box_constr_start}
		\State $k \gets 0$
		\Repeat
        	\Repeat
                \State $\Tilde{\thetab} \gets \Tilde{\thetab} + \eta \ \vb_{d}$ \Comment{Large step size $\eta$.}
        	\Until{$d(f_i(\Tilde{\thetab}), \data) > \epsilon$}
        	\State $\Tilde{\thetab} \gets \Tilde{\thetab} - \eta \ \vb_{d}$
        	\State $\eta \gets \eta/2$ \Comment{More accurate region boundary}
        	\State $k \gets k + 1$
    	\Until $k = K$
    	\State Set final $\Tilde{\thetab}$ as region end point. \label{algstep:box_constr_end}
    	\State Repeat steps~\ref{algstep:box_constr_start}~-~\ref{algstep:box_constr_end} for $\mathbf{v}_{d} = - \mathbf{v}_{d}$
	\EndFor
	\State Fit a rectangular box around the region end points and define $q_i$ as uniform distribution
	\end{algorithmic}
\end{algorithm}

% \begin{minipage}{0.46\textwidth}
% \begin{algorithm}[H]
%     \centering
%     \caption{Evaluate unnormalised posterior - Gradient approach. Requires $g_i(\theta), p(\theta)$}\label{alg:posterior_GB}
%     \begin{algorithmic}[1]
%       \State $k \leftarrow 0$
%         \For {$i \gets 1 \textrm{ to } n_1$}
%           \If {$g_i(\theta) > \epsilon$}
%             \State $k \leftarrow k + 1$
%           \EndIf
%           \EndFor
%       \Return{$kp(\theta)$}
%     \end{algorithmic}
% \end{algorithm}
% \end{minipage}
% \hfill
% \begin{minipage}{0.46\textwidth}
% \begin{algorithm}[H]
%     \centering
%     \caption{Evaluate unnormalised posterior - GP approach. Requires $\hat{d}_i(\theta), p(\theta)$}\label{alg:posterior_GP}
%     \begin{algorithmic}[1]
%       \State $k \leftarrow 0$
%         \For {$i \gets 1 \textrm{ to } n_1$}
%           \If {$d_i(\theta) > \epsilon$}
%             \State $k \leftarrow k + 1$
%           \EndIf
%           \EndFor
%       \Return{$kp(\theta)$}
%     \end{algorithmic}
% \end{algorithm}
% \end{minipage}

% \begin{minipage}{0.46\textwidth}
\begin{algorithm}[H]
    \centering
    \caption{Sampling. Requires a function of distance $(g_i(\theta)$ or $\hat{d}_i$ or $\hat{g}_i), p(\theta), q_i$}\label{alg:sampling_GB}
    \begin{algorithmic}[1]
      \For {$i \gets 1 \textrm{ to } n_1$}
      \For {$j \gets 1 \textrm{ to } n_2$}
          \State $\thetab_{ij} \sim q_i$
          \If {$g_i(\thetab_{ij}) > \epsilon$}
            \State Reject $\theta_{ij}$
          \Else {}
            \State $w_{ij} = \frac{p(\thetab_{ij})}{q(\thetab_{ij})}$
            \State Accept $\thetab_{ij}$, with weight $w_{ij}$
          \EndIf
      \EndFor
      \EndFor
    \end{algorithmic}
\end{algorithm}
% \end{minipage}
% \hfill
% \begin{minipage}{0.46\textwidth}
% \begin{algorithm}[H]
%     \centering
%     \caption{Sampling - GP approach. Requires $\hat{d}_i(\theta), p(\theta), q_i$}\label{alg:sampling_GP}
%     \begin{algorithmic}[1]
%       \For {$i \gets 1 \textrm{ to } n_1$}
%       \For {$j \gets 1 \textrm{ to } n_2$}
%           \State $\theta_{ij} \sim q_i$
%           \If {$\hat{d}_i(\theta_{ij}) > \epsilon$}
%             \State Reject $\theta_{ij}$
%           \Else {}
%             \State $w_{ij} = \frac{p(\theta_{ij})}{q(\theta_{ij})}$
%             \State Accept $\theta_{ij}$, with weight $w_{ij}$
%           \EndIf
%       \EndFor
%       \EndFor
%     \end{algorithmic}
% \end{algorithm}
% \end{minipage}

\subsection{Engine for Likelihood-Free Inference (ELFI) package}
\label{subsec:elfi}
The Engine for Likelihood-Free Inference (ELFI) \autocite{1708.00707}
is a Python package dedicated to Likelihood-Free Inference (LFI). ELFI
models in a convenient manner all the fundamental components of a
probabilistic model such as priors, simulators, summaries and
distances. Furthermore, ELFI already supports some recently proposed
likelihood-free inference methods.

\subsubsection{Modelling}
\label{sec:modelling}

\begin{figure}[!ht]
    \begin{center}
      \includegraphics[width=0.8\textwidth]{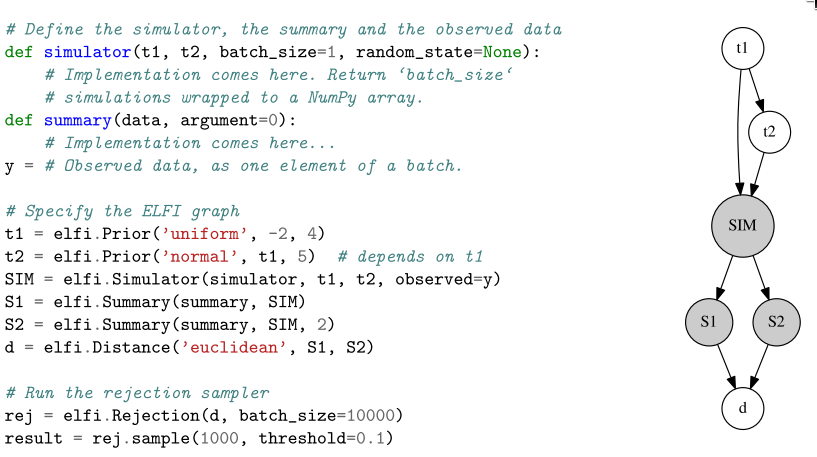}
    \end{center}
    \caption[Baseline example for creating an \pinline{ELFI} model]{Baseline example for creating an \pinline{ELFI} model. Image taken from \cite{1708.00707}}
    \label{fig:elfi}
\end{figure}

ELFI models the probabilistic model as a Directed Acyclic Graph (DAG);
it implements this functionality based on the package
\pinline{NetworkX}, which is designed for creating general purpose
graphs. Although not restricted to that, in most cases the structure
of a likelihood-free model follows the pattern presented in
figure~\ref{fig:elfi}; there are edges that connect the prior
distributions to the simulator, the simulator is connected to the
summary statistics that consequently are connected to the
distance. The distance is the output node. Samples can be obtained
from all nodes through sequential sampling. The nodes that are defined
as \pinline{elfi.Prior}\footnote{The \pinline{elfi.Prior} functionality
  is a wrapper around the \pinline{scipy.stats} package.} are automatically
considered as the parameters of interest and are the only nodes that,
apart from sampling, should also provide PDF evaluation. The function
passed as argument in the \pinline{elfi.Summary} node can be any valid
Python function. Finally, the observations should be passed in the
appropriate node through the argument \pinline{observed}.

\subsubsection{Inference Methods}
\label{sec:inference-methods}

The inference Methods implemented at the ELFI follow some common
guidelines;

\begin{itemize}
\item the initial argument should is the output node of the model. It
  is followed by the rest hyper-parameters of the method.
\item each inference method provides a central sampling
  functionality. In most cases it is named
  \pinline{<method_name>.sample()}.
\end{itemize}

The collection of likelihood-free inference methods implemented so far
contain the \textit{ABC Rejection Sampler}, the \textit{Sequential
  Monte Carlo ABC Sampler} and the \textit{Bayesian Optimisation for
  Likelihood-Free Inference (BOLFI)}. The latter has methodological
similarities to the ROMC method that we implement in the current work.

%%%%%%%%%%%%%%%%%%%%%%%%%%%%%%%%%%%%%%%% 
\clearpage
\section{Implementation}
In this section, we will exhibit the implementation of the ROMC
inference method in the ELFI package. The presentation is divided in
two logical blocks; In section~\ref{subsec:user} we present our method from the \emph{user's point-of-view}, i.e.\ what functionalities a practitioner can use for performing the inference in a simulator-based model. For showing the functions in-practice, we set-up a simple running example and
illustrate the functionalities on top of it. In section~\ref{subsec:developers} we delve into the internals of the
code, presenting the details of the implementation. This
section mainly refers to a \emph{researcher or a developer} who
would like to use ROMC as a meta-algorithm and experiment with novel
approaches. We have designed our implementation preserving
extensibility and customisation; hence, a researcher may intervene in
parts of the method without too much effort. A driver example which
demonstrates how to extend the method with custom utilities is
also contained in this section.

\subsection{Implementation description from the user's point-of-view}
\label{subsec:user}
In section \ref{subsec:general_design} we present an overview of the implementation. Afterwards, we divide the functionalities in 3 parts; in section \ref{subsec:training} we present the methods used for training (fitting) the model, in section \ref{subsec:inference} those used for performing the inference and in section \ref{subsec:evaluation} those used for evaluating the approximate posterior and the obtained samples. 

\subsubsection{General design} 
\label{subsec:general_design}
In figure \ref{fig:romc_overview} we present an overview of our
implementation; one may interpret figure~\ref{fig:romc_overview} as a
depiction of the main class of our implementation, called
\pinline{ROMC}, while the entities inside the green and blue
ellipses are the main functions of the class. Following Python's
naming principles, the methods starting with an underscore (green
ellipses) represent internal (private) functions and are not meant to
be used by a user, whereas the rest of the methods (blue ellipses) are
the functionalities the user interacts with. As mentioned before, the
implementation favours extensibility; the building blocks that compose
the method have been designed in a modular fashion so that a
practitioner may replace them without the method to collapse.

Figure \ref{fig:romc_overview} groups the ROMC implementation into the
training, the inference and the evaluation part. The training part includes all the
steps until the computation of the proposal regions; sampling the
nuisance variables, defining the optimisation problems, solving them,
constructing the regions and fitting local surrogate models. The
inference part comprises of evaluating the unnormalised posterior (and
the normalised one, in low-dimensional cases), sampling and computing
an expectation. Moreover, the ROMC implementation provides some
utilities for inspecting the training process, such as plotting the
histogram of the distances
$d^*_i = g_i(\theta_i^*), \: \forall i \in \{1, \ldots, n_1 \}$ after
solving the optimisation problems and visualising the constructed
bounding box\footnote{if the parametric space is up to $2D$}. Finally,
two functionalities for evaluating the inference are implemented; (a)
computing the Effective Sample Size (ESS) of the weighted samples and
(b) measuring the divergence between the approximate posterior the
ground-truth, if the latter is available.\footnote{Normally, the
  ground-truth posterior is not available; However, this functionality
  is useful in cases where the posterior can be computed numerically
  or with an alternative method (i.e.\ ABC Rejection Sampling) and we
  would like to measure the discrepancy between the two
  approximations.}
  
\subsubsection*{Parallelising the processes}

As stated above, the most critical advantage of the ROMC method is that it can be fully parallelised. In our implementation, we have exploited this fact by implementing a parallel version in the following tasks; (a) solving the optimisation problems, (b) constructing bounding box regions, (c) sampling and (d) evaluating the posterior. Parallelism has been achieved using the package \pinline{multiprocessing}. The specific package enables concurrency, using subprocesses instead of threads, hence side-stepping the Global Interpreter (GIL). In our implementation we use the \pinline{Pool} object, which offers a convenient means of parallelising the execution of a function across multiple input values (data parallelism). For activating the parallel version of the algorithms, the user has to just pass the argument \pinline{parallelize=True} at the initialisation of the \pinline{ROMC} method.

\subsubsection*{Simple one-dimensional example}

For illustrating the functionalities we choose as running example the
following model, introduced by~\autocite{Ikonomov2019},

\begin{gather} \label{eq:1D_example} p(\theta) =
\mathcal{U}(\theta;-2.5,2.5)\\ p(y|\theta) = \left\{
    \begin{array}{ll} \theta^4 + u & \mbox{if } \theta \in [-0.5, 0.5]
\\ |\theta| - c + u & \mbox{otherwise}
    \end{array} \right.\\ u \sim \mathcal{N}(0,1)
\end{gather}

\noindent

\begin{figure}[h]
    \begin{center}
\includegraphics[width=0.7\textwidth]{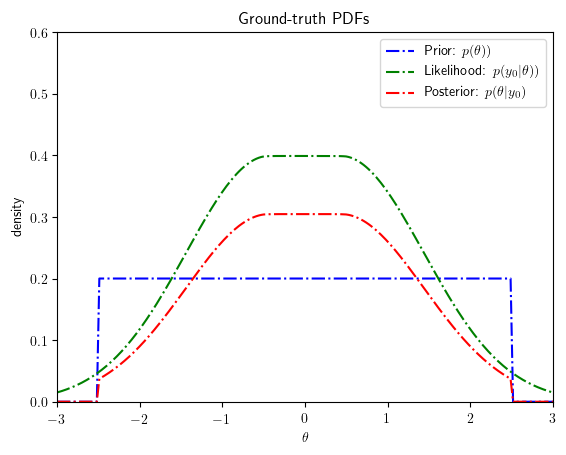}
    \end{center}
  \caption[Ground-truth posterior distribution of the simple 1D
example.]{Ground-truth posterior distribution of the simple 1D
example.}
  \label{fig:example_gt}
\end{figure}

In the model \eqref{eq:1D_example}, the prior is the uniform
distribution in the range $[-2.5, 2.5]$ and the likelihood a Gaussian
distribution. There is only one observation $y_0 = 0$. The inference
in this particular example can be performed quite easily, without
incorporating a likelihood-free inference approach. We can exploit this
fact for validating the accuracy of our implementation. The
ground-truth posterior, approximated computationally, is shown in
figure \ref{fig:example_gt}.

\subsubsection*{ELFI code for modelling the example}

In the following code snippet, we code the model at \textit{ELFI} and
we initialise the ROMC inference method. We observe that the
initialisation of the ROMC inference method is quite intuitive; we
just pass the final (distance) node of the simulator as argument, as
in all $\textit{ELFI}$ inference methods. The argument
\pinline{bounds}, although optional, is important for many
functionalities (e.g. approximating the partition function, setting
the bounds of the Bayesian optimisation etc.) so it is recommended to
be passed.

\begin{pythoncode}
  import elfi import scipy.stats as ss
  import numpy as np
  
  def simulator(t1, batch_size=1,random_state=None):
      if t1 < -0.5:
          y = ss.norm(loc=-t1-c, scale=1).rvs(random_state=random_state)
      elif t1 <= 0.5:
          y = ss.norm(loc=t1**4, scale=1).rvs(random_state=random_state)
      else:
          y = ss.norm(loc=t1-c, scale=1).rvs(random_state=random_state)
      return y

  # observation
  y = 0
      
  # Elfi graph
  t1 = elfi.Prior('uniform', -2.5, 5)
  sim = elfi.Simulator(simulator, t1, observed=y)
  d = elfi.Distance('euclidean', sim)

  # Initialise the ROMC inference method
  bounds = [(-2.5, 2.5)] # limits of the prior
  parallelize = True # activate parallel execution
  romc = elfi.ROMC(d, bounds=bounds, parallelize=parallelize)
\end{pythoncode}

\begin{figure}[!ht]
    \begin{center}
\includegraphics[width=0.8\textwidth]{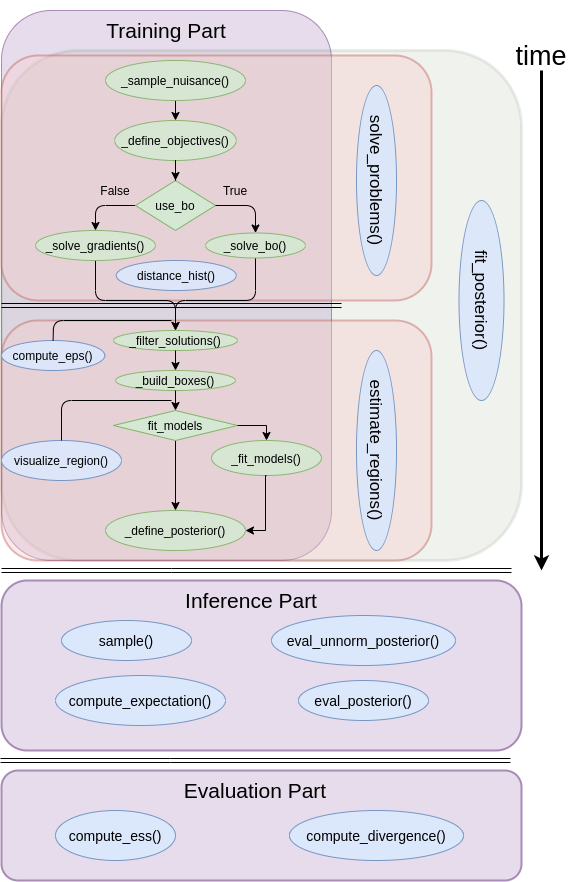}
    \end{center}
    \caption[Overview of the ROMC implementation.]{Overview of the
ROMC implementation. The training part follows a sequential pattern;
the functions in the green ellipses must be called in a sequential
fashion for completing the training part and define the posterior
distribution. The functions in blue ellipses are the functionalities
provided to the user.}
    \label{fig:romc_overview}
\end{figure}

\subsubsection{Training part} 
\label{subsec:training}
The training part contains the following 6 functionalities:

\begin{enumerate}[label=(\roman*)]
\item \mintinline{python}{romc.solve_problems(n1, use_bo=False, optimizer_args=None, seed=None)}
\item \mintinline{python}{romc.estimate_regions(eps_filter,}
  
      \mintinline{python}{                      use_surrogate=None, region_args=None,}
  
      \mintinline{python}{                      fit_models=False, fit_models_args=None,}
  
      \mintinline{python}{                      eps_region=None, eps_cutoff=None)}
      
    \item \mintinline{python}{romc.fit_posterior(n1, eps_filter, use_bo=False, optimizer_args=None,}
  
          \mintinline{python}{                   seed=None, use_surrogate=None, region_args=None,}
  
          \mintinline{python}{                   fit_models=False, fit_models_args=None,}
  
          \mintinline{python}{                   eps_region=None, eps_cutoff=None)}

\item \mintinline{python}{romc.distance_hist(savefig=False, **kwargs)}
\item \mintinline{python}{romc.visualize_region(i, savefig=False)}
\item \mintinline{python}{romc.compute_eps(quantile)}
\end{enumerate}

\subsubsection*{Function (i): Define and solve the optimisation problems}

\pinline{romc.solve_problems(n1, use_bo=False, optimizer_args=None, seed=None)}
\vspace{5mm}

\noindent
This routine is responsible for (a) drawing the nuisance variables,
(b) defining the optimisation problems and (c) solving them using either a
gradient-based optimiser or Bayesian optimisation. The aforementioned
tasks are done in a sequential fashion, as show in
figure~\ref{fig:romc_overview}. The definition of the optimisation
problems is performed by drawing $n_1$ integer numbers from a discrete
uniform distribution $u_i \sim \mathcal{U}\{1, 2^{32}-1\}$. Each
integer $u_i$ is the seed used in ELFI's random simulator. Hence from
an algorithmic point-of-view drawing the state of all random
variables $\vb_i$ as described in the previous chapter, traces back to
just setting the seed that initialises the state of the pseudo-random
generator, before asking a sample from the simulator.

Finally, passing an integer number as the argument \pinline{seed}
absorbs all the randomness of the optimisation part (e.g.\ drawing
initial points for the optimisation procedure), making the whole
process reproducible and deterministic.

Setting the argument \pinline{use_bo=True}, chooses the Bayesian
Optimisation scheme for obtaining $\thetab_i^*$. In this case, apart
from obtaining the optimal points $\thetab_i^*$, we also fit a Gaussian
Process (GP) as surrogate model $\hat{d}_i$. In the following steps,
$\hat{d}_i(\thetab)$ will replace $g_i(\thetab)$ when calling the
indicator function.

\subsubsection*{Function (ii): Construct bounding boxes and fit local surrogate models}

\mintinline{python}{romc.estimate_regions(eps_filter,}
  
      \mintinline{python}{                      use_surrogate=None, region_args=None,}
  
      \mintinline{python}{                      fit_models=False, fit_models_args=None,}
  
      \mintinline{python}{                      eps_region=None, eps_cutoff=None)}
\vspace{5mm}

This routine constructs the bounding boxes around the optimal points
$\thetab_i^* : i = 1, 2, \ldots, n_1$ following
Algorithm~\ref{alg:region_construction}. The Hessian matrix is
approximated based on the Jacobian $\hess_i = \jac_i^T \jac_i$. The
eigenvectores are computed using the function
\pinline{numpy.linalg.eig()} that calls, under the hood, the
\pinline{_geev LAPACK}. A check is performed so that the matrix
$\hess_i$ is not singular; if this is the case, the eigenvectors are
set to be the vectors of the standard Euclidean basis i.e.\
$\{ \mathbf{e_1} = (1, 0, \ldots), \mathbf{e_2} = (0,1,0,\ldots),
\text{etc} \}$. Afterwards, the limits are obtained by repeteadly querying
the distance function ($g_i(\thetab)$ or $\hat{d}(\thetab)$) along the
search directions. In section \ref{subsec:developers}, we provide some
details regarding the way the bounding box is defined as a class and
sampling is performed on it.

\subsubsection*{Function (iii): Perform all training steps in a single call}

\mintinline{python}{romc.fit_posterior(n1, eps_filter, use_bo=False, optimizer_args=None,}
  
          \mintinline{python}{                seed=None, use_surrogate=None, region_args=None,}
  
          \mintinline{python}{                fit_models=False, fit_models_args=None,}
  
          \mintinline{python}{                eps_region=None, eps_cutoff=None)}

\vspace{5mm}
\noindent

This function merges all steps for constructing the bounding box into
a single command. If the user doesn't want to manually inspect the
histogram of the distances before deciding where to set the threshold
$\epsilon$, he may call \pinline{romc.fit_posterior()} and the whole
training part will be done end-to-end. There are two alternatives for
setting the threshold $\epsilon$; the first is to set to a specific
value blindly and the second is to set at as a specific quantile of
the histogram of distances. In the second scenario the
\pinline{quantile} argument must be set to a floating number in the
range $[0,1]$ and \pinline{eps='auto'}.

\subsubsection*{Function (iv): Plot the histogramm of the optimal points}

\pinline{romc.distance_hist(**kwargs)}
\vspace{5mm}
\noindent

This function can serve as an intermediate step of manual inspection,
for helping the user choose which threshold $\epsilon$ to use. It
plots a histogram of the distances at the optimal point
$g_i(\thetab_i^*) : \{i = 1, 2, \ldots, n_1\}$ or
$d_i^*$ in case \pinline{use_bo=True}. The function accepts all
keyword arguments and forwards them to the underlying
\pinline{matplotlib.hist()} function; in this way the user may
customise some properties of the histogram, such as the number of bins
or the range of values.

\subsubsection*{Function (v): Plot the acceptance region of the objective functions}

\pinline{romc.visualize_region(i)}
\vspace{5mm}
\noindent

It can be used as an inspection utility for cases where the parametric
space is up to two dimensional. The argument $\pinline{i}$ is the
index of the corresponding optimization problem i.e.\ $i<n_1$.

\subsubsection*{Function (vi): Compute $\epsilon$ automatically based on the distribution of $d^*$}

\pinline{romc.compute_eps(quantile)}
\vspace{5mm}
\noindent

\noindent
This function return the appropriate distance value $d_{i=\kappa}^*$
where $\kappa = \lfloor \frac{quantile}{n} \rfloor$ from the
collection $\{ d_i^* \} \forall i = \{1, \ldots, n\}$ where $n$ is the
number of accepted solutions. It can be used to automate the selection
of the threshold $\epsilon$, e.g.\
\pinline{eps=romc.compute_eps(quantile=0.9)}.

\subsubsection*{Example}

Here we will illustrate the aforementioned functionalities using the
simple 1D example we set up in the previous section. The following
code snippet performs the training part in ELFI.

\begin{pythoncode}
  n1 = 500 # number of optimisation problems
  seed = 21 # seed for solving the optimisation problems
  eps = .75 # threshold for bounding box
  use_bo = False # set to True for switching to Bayesian optimisation

  # Training step-by-step
  romc.solve_problems(n1=n1, seed=seed, use_bo=use_bo)
  romc.theta_hist(bins=100)
  romc.estimate_regions(eps=eps)
  romc.visualize_region(i=1)

  # Equivalent one-line command
  # romc.fit_posterior(n1=n1, eps=eps, use_bo=use_bo, seed=seed)
\end{pythoncode}

\begin{figure}[h]
    \begin{center}
      \includegraphics[width=0.48\textwidth]{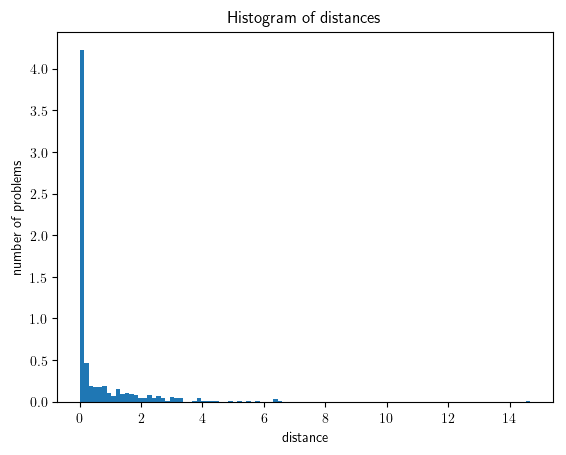}
      \includegraphics[width=0.48\textwidth]{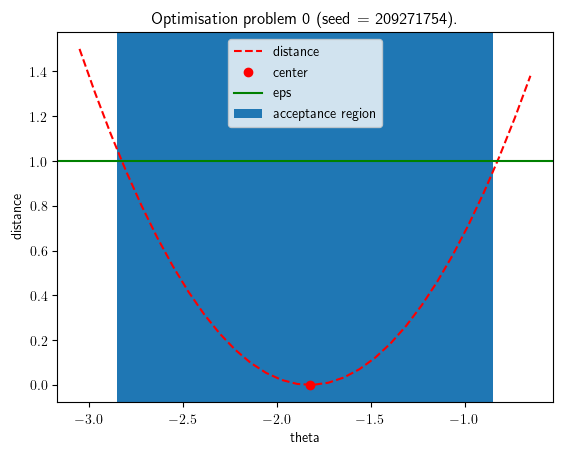}
    \end{center}
  \caption[Histogram of distances at the 1D example.]{Histogram of distances and visualisation of a specific region.}
  \label{fig:example_training_hist}
\end{figure}

As stated before, switching to the Bayesian optimisation scheme needs
nothing more the setting the argument \pinline{use_bo=True}; all the
following command remain unchanged. In figure
\ref{fig:example_training_hist} we illustrate the distribution of the
distances obtained and the acceptance area of the first optimisation
problem. We observe that most optimal points produce almost zero
distance.

\subsubsection{Inference part} 
\label{subsec:inference}
The inference part contains the 4 following functionalities:

\begin{enumerate}[label=(\roman*)]
\item \mintinline{python}{romc.sample(n2, seed=None)}
\item \mintinline{python}{romc.compute_expectation(h)}  
\item \mintinline{python}{romc.eval_unnorm_posterior(theta)}
\item \mintinline{python}{romc.eval_posterior(theta)}
\end{enumerate}

\subsubsection*{Function (i): Perform weighted sampling}

\mintinline{python}{romc.sample(n2)}
\vspace{5mm}

\noindent
This is the basic inference utility of the ROMC implementation; we
draw $n_2$ samples for each bounding box region. This gives a total of
$k \times n_2$, where $k < n_1$ is the number of the optimal points
remained after filtering\footnote{From the $n_1$ optimisation
  problems, only the ones with $g_i(\thetab_*) < \epsilon$ are kept
  for building a bounding box}. The samples are drawn from a uniform
distribution $q_i$ defined over the corresponding bounding box and the
weight $w_i$ is computed as in equation~\eqref{eq:sampling}. The
function stores an \pinline{elfi.Result} object as
\pinline{romc.result} attribute. The \pinline{elfi.Result} provides
some usefull functionalities for inspecting the obtained samples e.g.\
\pinline{romc.result.summary()} prints the number of the obtained
samples and their mean. A complete overview of these functionalities is
provided in ELFI's
\href{https://elfi.readthedocs.io/en/latest/api.html#elfi.methods.results.Sample}{official
  documentation}.

\subsubsection*{Function (ii): Compute an expectation}

\mintinline{python}{romc.compute_expectation(h)}
\vspace{5mm}

\noindent
This function computes the expectation
$E_{p(\thetab|\data)}[h(\thetab)]$ using
expression~\eqref{eq:expectation}. The argument \pinline{h} can be
any python \pinline{Callable}.

\subsubsection*{Function (iii): Evaluate the unnormalised posterior}
\mintinline{python}{romc.eval_unorm_posterior(theta, eps_cutoff=False)}
\vspace{5mm}

\noindent
This function computes the unnormalised posterior approximation using
expression~\eqref{eq:approx_posterior}.

\subsubsection*{Function (iv): Evaluate the normalised posterior}
\mintinline{python}{romc.eval_posterior(theta, eps_cutoff=False)}
\vspace{5mm}

\noindent
This function evaluates the normalised posterior. For doing so it
needs to approximate the partition function
$Z = \int p_{d,\epsilon}(\thetab|\data)d\thetab$; this is done using
the Riemann integral approximation. Unfortunately, the Riemann
approximation does not scale well in high-dimensional spaces, hence
the approximation is tractable only at low-dimensional parametric
spaces. Given that this functionality is particularly useful for
plotting the posterior, we could say that it is meaningful to be used
for up to $3D$ parametric spaces, even though it is not restricted to
that. Finally, for this functionality to work, the \pinline{bounds}
arguments must have been set at the initialisation of the
\pinline{elfi.ROMC} object.\footnote{The argument \pinline{bounds}
  should define a bounding box \emph{containing} all the mass of the
  prior; it may also contain redundant areas. For example, if the
  prior is the uniform defined over a unit circle i.e.\
  $c=(0,0), r=1$, the best bounds arguments is
  \pinline{bounds=[(-1,1),(-1,1)]}. However, any argument
  \pinline{bounds=[(-t,t),(-t,t)]} where $t\geq1$ is technically
  correct.}

\subsubsection*{Example - Sampling and compute expectation}

\begin{figure}[h]
    \begin{center}
      \includegraphics[width=0.48\textwidth]{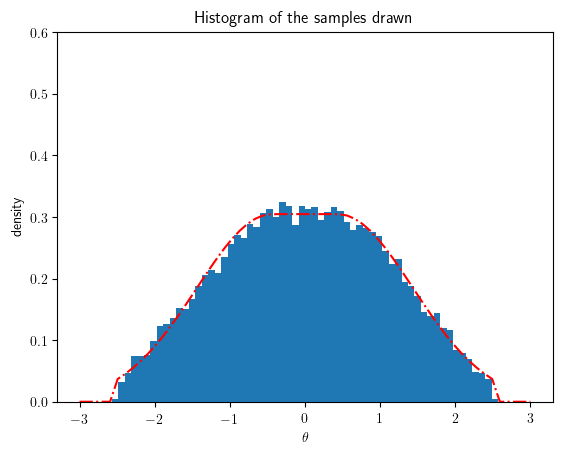}
      \includegraphics[width=0.48\textwidth]{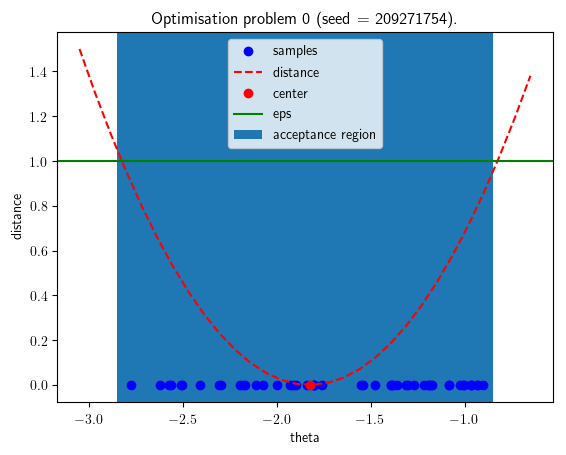}
    \end{center}
  \caption[Histogram of the obtained samples at the 1D example.]{(a) Left: Histogram of the obtained samples. (b) Right: Acceptance region around $\theta_1^*$ with the obtained samples plotted inside.}
  \label{fig:example_sampling}
\end{figure}

With the following code snippet, we perform weighted sampling from the
ROMC approximate posterior. Afterwards, we used some ELFI's built-in
tools to get a summary of the obtained samples. In figure
\ref{fig:example_sampling}, we observe the histogram of the weighted
samples and the acceptance region of the first deterministic function
(as before) alongside with the obtained samples obtained from
it. Finally, in the code snippet we demonstrate how to use the
\pinline{compute_expectation} function; in the current example we
define \pinline{h} in order to compute firstly the empirical mean and
afterwards the empirical variance. In both cases, the empirical result
is close to the ground truth $\mu = 0$ and $\sigma^2 = 1$.

\begin{pythoncode}
  seed = 21
  n2 = 50
  romc.sample(n2=n2, seed=seed)

  # visualize region, adding the samples now
  romc.visualize_region(i=1)

  # Visualise marginal (built-in ELFI tool)
  romc.result.plot_marginals(weights=romc.result.weights, bins=100, range=(-3,3))

  # Summarize the samples (built-in ELFI tool)
  romc.result.summary()
  # Number of samples: 1720
  # Sample means: theta: -0.0792

  # compute expectation
  print("Expected value   : %.3f" % romc.compute_expectation(h=lambda x: np.squeeze(x)))
  # Expected value   : -0.079

  print("Expected variance: %.3f" % romc.compute_expectation(h=lambda x: np.squeeze(x)**2))
  # Expected variance: 1.061
\end{pythoncode}

\subsubsection*{Example - Evaluate Posterior}

\begin{figure}[ht]
    \begin{center}
      \includegraphics[width=0.75\textwidth]{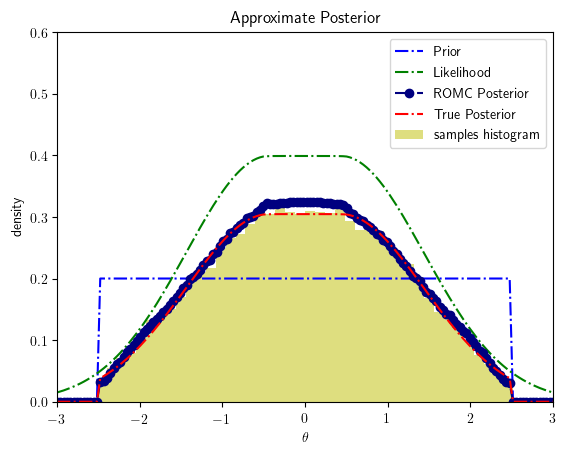}
    \end{center}
  \caption[Approximate posterior evaluation, at the 1D example.]{Approximate posterior evaluation.}
  \label{fig:approx_posterior}
\end{figure}

The \pinline{romc.eval_unnorm_posterior(theta)} evaluates the
posterior at point $\theta$ using expression
\eqref{eq:aprox_posterior}. The \pinline{romc.eval_posterior(theta)}
approximates the partition function
$Z = \int_{\thetab} p_{d,\epsilon}(\thetab|\data) d\thetab$ using the
Riemann approximation as explained above. In our simple example, this
utility can provide a nice plot of the approximate posterior as
illustrated in figure~\ref{fig:approx_posterior}. We observe that the
approximation is quite close to the ground-truth posterior.

\subsubsection{Evaluation part} 
\label{subsec:evaluation}
The ROMC implementation provides two functions for evaluating the inference results,

\begin{enumerate}[label=(\roman*)]
\item \mintinline{python}{romc.compute_divergence(gt_posterior, bounds=None, step=0.1,}

      \mintinline{python}{                        distance="Jensen-Shannon")}

\item \mintinline{python}{romc.compute_ess()}
\end{enumerate}

\subsubsection*{Function (i): Compute the divergence between ROMC approximation and a ground-truth posterior}

\pinline{romc.compute_divergence(gt_posterior, bounds=None, step=0.1,}

\pinline{                     distance="Jensen-Shannon")}
\vspace{5mm}

\noindent
This function computes the divergence between the ROMC approximation
and the ground truth posterior. Since the computation is performed
using the Riemann's approximation, this method can only work in low
dimensional parameter spaces; it is suggested to be used for up to the
three-dimensional parameter space. As mentioned in the beginning of
this chapter, in a real-case scenario it is not expected the
ground-truth posterior to be available. However, in cases where the
posterior can be approximated decently well with a numerical approach
(as in the current example) or with some other inference approach,
this function can provide a numerical measure of the agreement
between the two approximations. The argument \pinline{step} defines
the step used in the Riemann's approximation and the argument
\pinline{distance} can take either the \pinline{Jensen-Shannon} or the
\pinline{KL-divergence} value, for computing the appropriate distance.

\subsubsection*{Function (ii): Compute the effective sample size of the weighted samples}

\mintinline{python}{romc.compute_ess()}
\vspace{5mm}

\noindent
This function computes the Effective Sample Size (ESS) using the
following expression~\autocite{Sudman1967},

\begin{equation} \label{eq:ESS}
  ESS = \frac{(\sum_i w_i)^2}{\sum_i w_i^2}
\end{equation}

The ESS is a useful measure of the \textbf{actual} sample size, when
the samples are weighted. For example if in a population of $100$
samples one has a very large weight (e.g.\ $\approx 100$) whereas the
rest have small (i.e.\ $\approx 1$), the real sample size is close to
1; one sample dominates over the rest. Hence, the ESS provides a
measure of the equivalent uniformly weighted sample population.

\newpage

\begin{pythoncode}
res = romc.compute_divergence(wrapper, distance="Jensen-Shannon")                                 
print("Jensen-Shannon divergence: %.3f" % res)
# Jensen-Shannon divergence: 0.025

print("Nof Samples: %d, ESS: %.3f" % (len(romc.result.weights), romc.compute_ess()))
# Nof Samples: 19950, ESS: 16694.816
\end{pythoncode}

\subsection{Implementation details for the developer}
\label{subsec:developers}
In the previous sections we described ROMC from the user's
point-of-view; the demonstration focused on performing the inference
with the ready-to-use tools. Apart from this use-case, a practitioner
can also use ROMC as a meta-algorithm, adding custom algorithms as
part of the method. Our implementation allows such intervention. In
the rest of the chapter, we will initially present the main internal
classes and then then we will exhibit how to add custom methods.

\subsubsection{Entities presentation}

\begin{figure}[ht]
    \begin{center}
      \includegraphics[width=\textwidth]{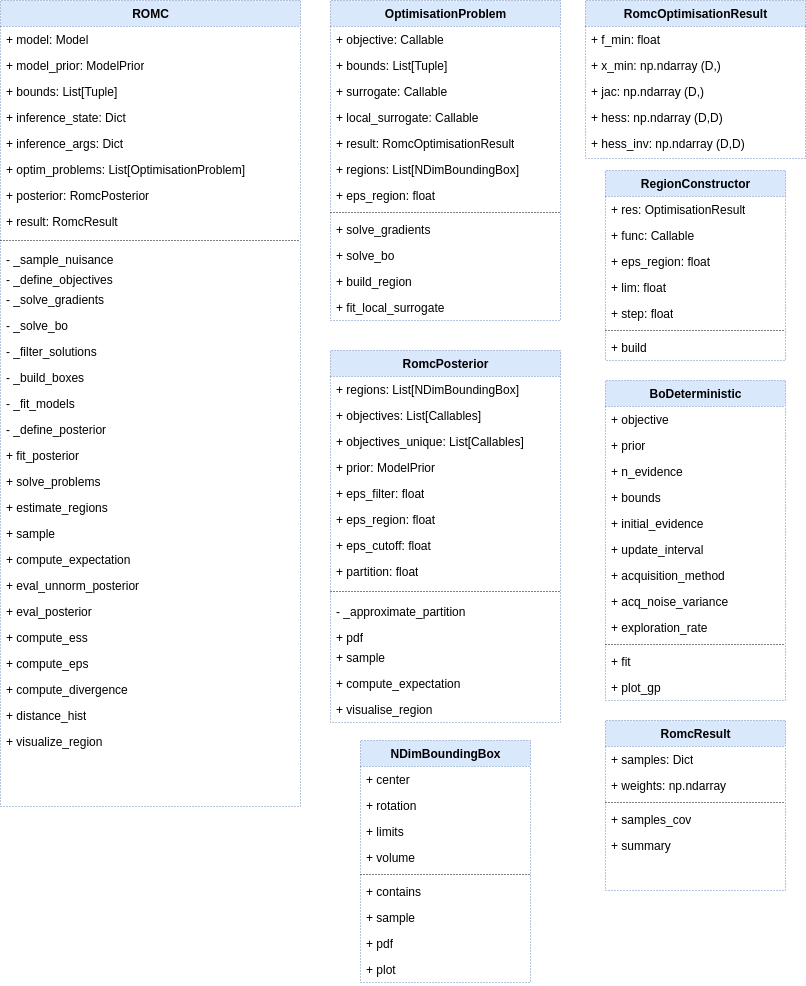}
    \end{center}
  \caption[Overview of the entities (classes) of the ROMC implementation.]{Overview of the entities (classes) used at the deployment of the ROMC inference method, at the \pinline{ELFI} package.}
  \label{fig:example_training}
\end{figure}

In figure \ref{fig:example_training} we provide an overview of the
classes used in the algorithm. The class $\pinline{ROMC}$ can be
thought as the interface between the user and the method. The rest of
the classes are responsible for the back-end functionality. It is
important for the reader to remember the sequence of the events for
performing the inference, as demonstrated in figure
\ref{fig:romc_overview}.

The class \pinline{ROMC} is the main class of the method; it is
initialised when the user calls the method for the first time and its
state is updated throughout the inference.The initialisation of the
\pinline{ROMC} object sets the attributes \pinline{model},
\pinline{model_prior}, \pinline{bounds} and \pinline{inference_state} to
the appropriate values; the rest of the attributes are set to
\pinline{None}.\footnote{in all cases, we use the value None for
  indicating that an attribute is not yet initialised."}.

The \pinline{_sample_nuisance()} routine samples $n_1$ integers from
the discrete integer distribution. The \pinline{_define_objectives()}
routine initialises $n_1$ \pinline{OptimisationProblem} objects,
appends them to a \pinline{List} and stores this \pinline{List} as the
\pinline{romc.optim_problems} attribute. The
\pinline{_define_objectives()} function, initialises only the
attributes \pinline{objective} and \pinline{bounds} of the
\pinline{OptimisationProblem}; the rest are set to \pinline{None}. The
attribute \pinline{objective}, which is a \pinline{Callable}, is
created by setting the seed of the \pinline{model.generate()} to a
specific integer, turning the random generator to a deterministic.

Afterwards, depending on the boolean argument
\pinline{use_bo=True/False}, the function \linebreak \pinline{_solve_gradients}
or \pinline{_solve_bo} is called. Both of them follow the same
sequence of steps; for each optimisation problem they (a) solve it,
(b) create a \pinline{RomcOptimisationResult} object with the solution
and (c) store it as the \pinline{OptimisationProblem.result}
attribute. Although they both apply the same sequence of steps, each
method uses a different optimisation scheme;
\pinline{_solve_gradients} uses the \pinline{minimize} class of the
\pinline{scipy.optimize} library \autocite{2020SciPy-NMeth}, whereas
\pinline{_solve_bo} uses the \pinline{BoDeterministic} class.
\pinline{BoDeterministic} is a class we implemented for performing
Bayesian Optimisation and fitting a Gaussian Process surrogate model
to the objective function. It relies on the Gpy framework
\autocite{gpy2014} for fitting the Gaussian Process. The only
difference between \pinline{_solve_gradients} and \pinline{_solve_bo},
as effect in the state of the \pinline{OptimisationProblem} class, is
the initialisation of the attribute 
\pinline{OptimisationProblem.surrogate}, which is done only by
\pinline{_solve_bo}. The attribute is set to a \pinline{Callable} that
wraps the \pinline{GPyRegression.predict_mean} function. If
\pinline{_solve_gradients} is called, the attribute remains
\pinline{None}.

At this level, all optimisation problems have been solved. The method
\pinline{_filter_solutions} is used for discarding the optimisation
results that are over threshold. Afterwards, \pinline{_build_boxes}
estimates the bounding boxes around the accepted objective
functions. For each accepted objective function, a
\pinline{RegionConstructor} object is initialised in order to
construct the region. In the current implementation, for each
objective function we construct a single bounding box, but this may
change in a future approach; for being able to support multiple
bounding boxes per objective function, we decided to return a
\pinline{List} of \pinline{NDimBoundingBox} objects\footnote{Each
  \pinline{NDimBoundingBox} object represents a region.}. The
\pinline{List} is stored as the \pinline{OptimisationProblem.regions}
attribute.

By now, we have estimated the bounding boxes around the optimal
points. The last step, before defining the posterior, is the fitting
local surrogate models. This step is optional and is performed only
if the argument \pinline{fit_models} is set to \pinline{True}. The
routine \pinline{_fit_models}, fits a quadratic model on the area
around the optimal point for each objective function. For achieving
so, it asks for samples from the \pinline{NDimBoundingBox} object and
evaluates them using the \pinline{OptimisationProblem.objective}
function. Afterwards, based on these points, it fits a quadratic model
using the \pinline{linear_model.LinearRegression} and
\pinline{preprocessing.PolynomialFeatures} functions of the
\pinline{scikit-learn} package \autocite{scikit-learn}. The trained
model is stored as the attribute
\pinline{OptimisationProblem.local_surrogate}.

Finally, the \pinline{_define_posterior} method is used for creating
the \pinline{RomcPosterior} object and storing it as the
\pinline{ROMC.posterior} attribute. The method collects (a) all the
bounding boxes created so far (accessing the
\pinline{OptimisationProblem.regions} attributes and (b) all the objective functions. As the objective function it collects either \pinline{OptimisationProblem.local_surrogate} or \linebreak \pinline{OptimisationProblem.surrogate} or \pinline{OptimisationProblem.objective}, depending on which one is available based on the previous steps.

The description above summarises the sequence of steps needed for the
training part of the method. The conclusion is that a \pinline{ROMC}
object is initialised when the user calls the method. Throughout the
inference process, the \pinline{ROMC} object is always as a specific state,
which gets updated whenever an algorithmic step is executed. The rest
of the classes provide objects which are stored as attributes of
\pinline{ROMC}.

\subsubsection{Extensibility of the ROMC method}

In this section we will explain how a practitioner may replace some
parts of the ROMC method with their custom alogorithms. We can locate
at least four such points; (a) the gradient-based solver (b) the
Bayesian Optimisation solver (c) the bounding box region construction
and (d) the fitting of the local surrogate models. Each of the
aforementioned tasks may be approached with completely different
algorithms than the ones we propose, without the rest of the method to
change.

The four replacable parts described above, are solved using the four
methods of the \linebreak \pinline{OptimisatioProblem} class;
\pinline{solve_gradients(**kwargs)}, \pinline{solve_bo(**kwargs)}, \linebreak
\pinline{build_region(**kwargs)},
\pinline{fit_local_surrogate(**kwargs)}. Therefore, the practitioner
should not alter at all the basic \pinline{ROMC} class. Instead, they
should deploy a custom optimisation problem class which inherits the
basic \pinline{OptimisatioProblem} class and overwrites the above four
functiones with custom ones. The only rule that should be followed
carefully, is that the cutom methods must have the same effect on the
attributes of the \pinline{ROMC} class, i.e. update them with the
appropriate objects as presented in table \ref{tab:extensibility}. For
example, a function that replaces \pinline{solve_gradients()} must
have the effect of storing a \pinline{RomcOptimisationResult} object
to the \pinline{OptimisationObject} attribute.

\begin{center} \label{tab:extensibility} \captionof{table}[Demonstration of the appropriate effect of each replaceable routine]{Table
    explaining the appropriate effect of each replaceable routine,
    i.e. which object they should attach to the appropriate
    attribute. The functions of the first column (\pinline{ROMC}
    class) call the correpsonding functions of the second column
    (\pinline{OptimisationProblem} class). The functions of the second
    column should execute their main functionality and update the
    appropriate attribute with a certain object, as described in the
    third column. }

\begin{tabular}{ c|c }
\hline
\pinline{OptimisationProblem} & \pinline{Effect} \\
\hline \hline
\pinline{solve_gradients()} & \pinline{result <- RomcOptimisationResult} \\
\hline
\multirow{ 2}{*}{\pinline{solve_bo()}} & \pinline{result <- RomcOptimisationResult} \\
& \pinline{surrogate <- Callable} \\
\hline
\pinline{build_region()} & \pinline{regions <- List[NDimBoundingBox]}\\
\hline
\pinline{fit_local_surrogate()} & \pinline{local_surrogate <- Callable}\\
\hline
\end{tabular}
\end{center}

\subsubsection*{Example: use a Neural Network as a local surrogate model}

Let's say we have observed that the local area around $\theta_i^*$ is
too complex to be represented by a simple quadratic model\footnote{as
  in the current implementation}. Hence, the user selects a neural
network as a good alternative. In the following snippet, we
demonstrate how they could implement this enhancement, without much
effort; (a) they have to develop the neural network using the package
of their choice (b) they must create a custom optimisation class which
inherites the basic \pinline{OptimisationClass} and (c) they have to
overwrite the \pinline{fit_local_surrogate} routine, with one that
sets the neural network's prediction function as the
\pinline{local_surrogate} attribute. The argument \pinline{**kwargs}
may be used for passing all the important arguments e.g.\ training
epochs, gradient step etc. If, for example, they would like to set the
size of the training set dynamically, we may replace \pinline{x =
  self.regions[0].sample(30)} with \pinline{x =
  self.regions[0].sample(kwargs["nof_examples"])}. Finally, they must
pass the custom optimisation class, when calling the \pinline{ROMC}
method.

\newpage 

\begin{pythoncode}
  class NeuralNetwork:
      def __init__(self, **kwargs):
          # set the input arguments

      def train(x, y):
          # training code

      def predict(x):
          # prediction code

  # Inherit the base optimisation class
  class customOptim(elfi.methods.parameter_inference.OptimisationProblem):
      def __init__(self, **kwargs):
          super(customOptim, self).__init__(**kwargs)

      # overwrite the function you want to replace
      def fit_local_surrogate(**kwargs):
          # init and train the NN
          x = self.regions[0].sample(30) # 30 training points
          y = [np.array([self.objective(ii) for ii in x])]
          nn = NeuralNet()
          nn.train(x,y)

          # set the appropriate attribute
          self.local_surrogate = nn.predict

          # update the state
          self.state["local_surrogate"] = True

  # pass the custom inference method as argument
  romc = elfi.ROMC(dist, bounds, custom_optim_class=customOptim)
\end{pythoncode}

\clearpage

%%%%%%%%%%%%%%%%%%%%%%%%%%%%%%%%%%%%%%%%
\section{Experiments}
This section incorporates two real-case examples for demonstrating the
accuracy of the method both at a conceptual and the implementation
level. The first example has tractable likelihood, therefore the
ground-truth information is available and can be used for
validation. The second example is the second-degree moving average
model, which is a basic example used by the ELFI package for checking
the inference methods. This model is also chosen to illustrate that
our implementation performs well at a general model, not implemented
by us.

\subsection{Example 1: Simple 2D example}
\label{subsec:ex1}
This examples is implemented for validating that the ROMC
implementation works accurately in a multidimensional parameter
space.

\subsubsection*{Problem Definition}

The equations describing this inference problem are presented below.

\begin{gather} \label{eq:ex1_equations}
  p(\thetab) = p(\theta_1)p(\theta_2)
  = \mathcal{U}(\theta_1; -2.5, 2.5) \mathcal{U}(\theta_2; -2.5, 2.5)\\
  p(\yb|\thetab) = \mathcal{N}(\yb;\thetab, \mathcal{I})\\
  p(\thetab|\yb) = \frac{1}{Z}p(\thetab)p(\yb|\thetab)\\
  Z = \int_{\thetab} p(\thetab)p(\yb|\thetab) d\thetab
\end{gather}

\noindent
In this simple example, it is feasible to evaluate the likelihood and
the unnormalised posterior. Additionally, the partition function $Z$
can be estimated by numerical integration (i.e. using Riemann's
approximation). Hence, it is feasible to compute the ground-truth
posterior with numerical approximation. Setting the observation
$\data$ to $(-0.5,0.5)$, the ground-truth posterior is illustrated in
figure \ref{fig:ex2_4}. In table \ref{tab:ex2}, we present the
ground-truth statistics i.e.\ $\mu, \sigma$ of the marginal posterior
distributions.

\subsubsection*{Performing the inference}

\begin{figure}[ht]
    \begin{center}
      \includegraphics[width=0.48\textwidth]{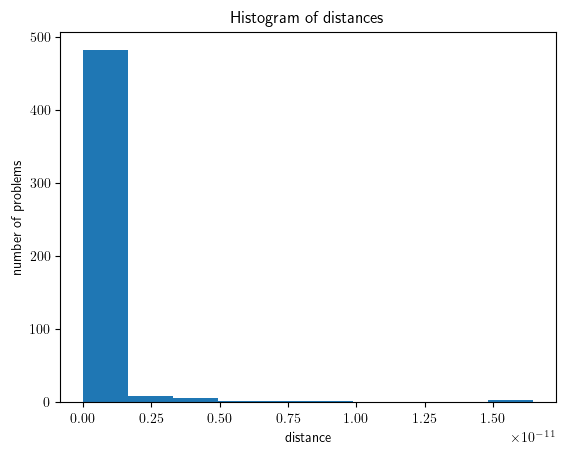}
      \includegraphics[width=0.48\textwidth]{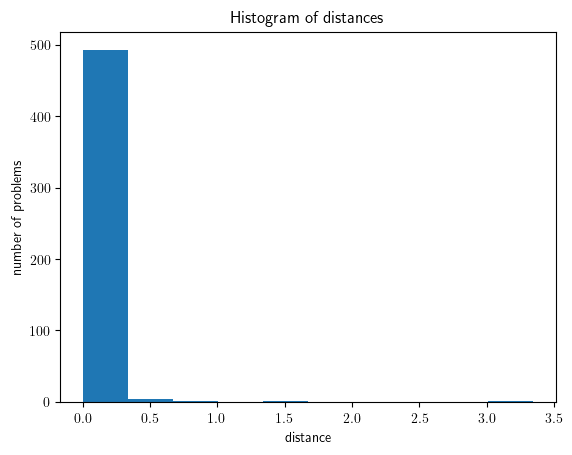}
    \end{center}
    \caption[2D example, histogram of distances]{Histogram of distances $d_i^*, i \in \{1, \ldots,
      n_1\}$. The left graph corresponds to the gradient-based
      approach and the right one to the Bayesian optimisation
      approach.}
  \label{fig:ex2_1}
\end{figure}

\begin{figure}[ht]
    \begin{center}
      \includegraphics[width=0.48\textwidth]{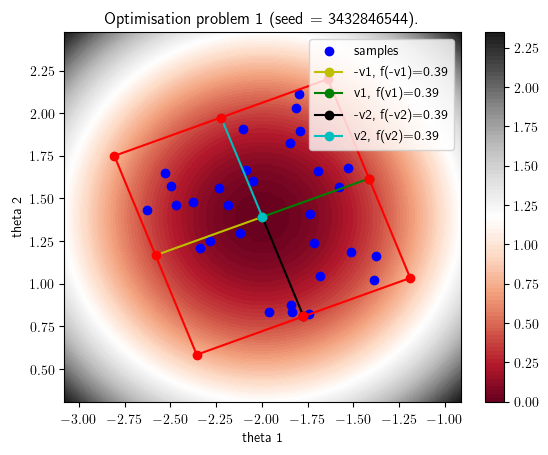}
      \includegraphics[width=0.48\textwidth]{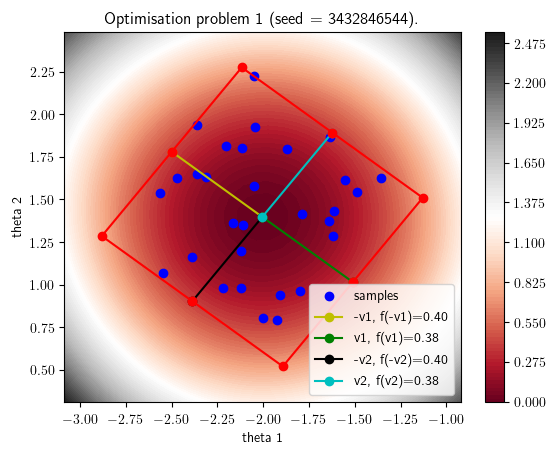}
    \end{center}
    \caption[2D example, the bounding box of the $1^{st}$ optimisation problem.]{The acceptance region and the bounding box of the
      $1^{st}$ optimisation problem; In the left plot, it is
      constructed with a gradient-based approach and in the right
      figure with Bayesian optimisation. We observe that both
      approaches construct a similar bounding box.}
  \label{fig:ex2_2}
\end{figure}

We perform the inference using the following hyperparameters
$n_1=500, n_2=30, \epsilon=0.4$. This set-up leads to a total of
$15000$ samples. As observed in the histogram of distances (figure
\ref{fig:ex2_1}), in the gradient-based approach, all optimisation
problems reach an almost zero-distance end point; hence all optimal
points are accepted. In the Bayesian optimisation scheme, the vast
majority of the optimisation procedures has the same behaviour; there
are only 4 optimal distances above the limit. In figure
\ref{fig:ex2_2}, the acceptance area of a specific optimisation
problem is demonstrated. We observe that both optimisation schemes
lead to a similar bounding box construction. This specific example is
representative of the rest of the optimisation problems; due to the
simplicity of the objective function, in most cases the optimal points
are similar and the surrogate model represents accurately the local
region. Hence, similar proposal regions are obtained by the two
optimisation alternatives.

\begin{figure}[ht]
    \begin{center}
      \includegraphics[width=0.48\textwidth]{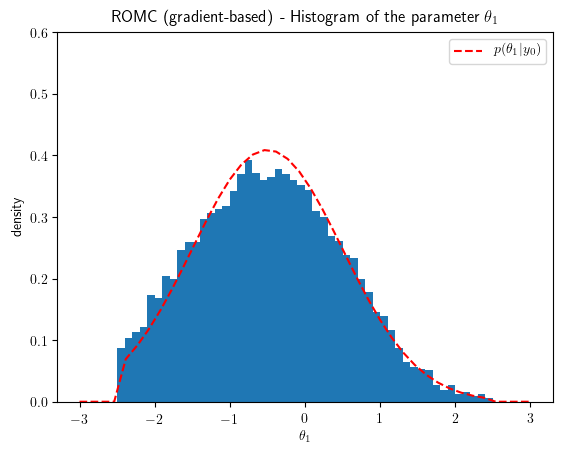}
      \includegraphics[width=0.48\textwidth]{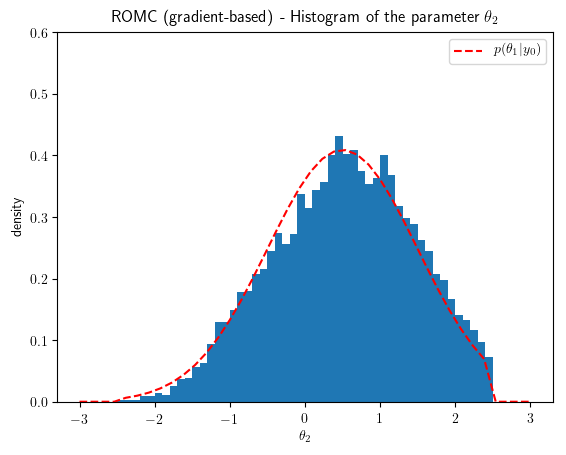}\\
      \includegraphics[width=0.48\textwidth]{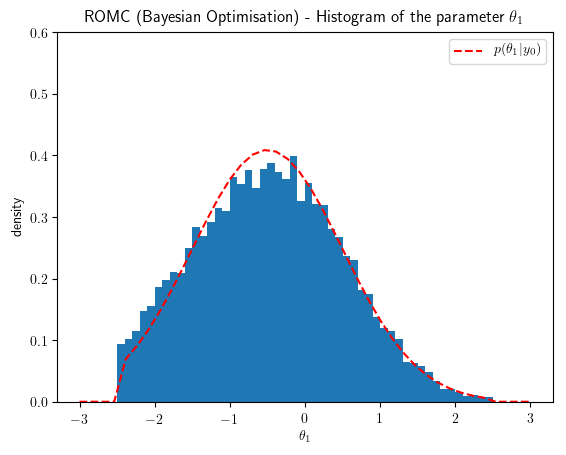}
      \includegraphics[width=0.48\textwidth]{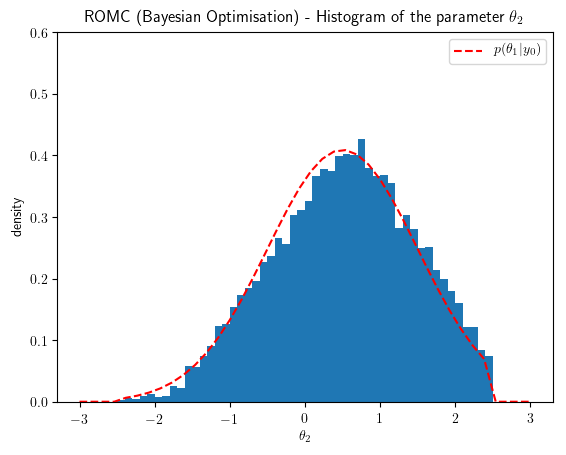}\\
    \end{center}
    \caption[2D example, histogram of the weighted samples.]{Histogram of the weighted samples. The first row of plots
      refers to the gradient-based optimisation scheme, while the
      second row to the Bayesian optimisation. We represent the
      ground-truth marginal distribution $p(\theta_i|\data$ with the
      red dotted line. We observe that the samples follow the ground
      truth distribution quite accurately.}
  \label{fig:ex2_3}
\end{figure}

The histograms of the marginal distributions, based on the weighted
samples, are presented in figure \ref{fig:ex2_3}. In the same figure,
we also plot the ground-truth distribution with the red dotted
line. We observe that the weighted samples follow quite accurately the
ground-truth distribution. This is also confirmed by the statistics
provided in table \ref{tab:ex2}; the sample mean $\mu$ and standard
deviation $\sigma$ are similar to the ground-truth for both parameters
$\theta_1$ and $\theta_2$. We also observe that both optimisation
schemes produce accurate samples.

\begin{figure}[ht]
  \begin{center}
      \includegraphics[width=0.5\textwidth]{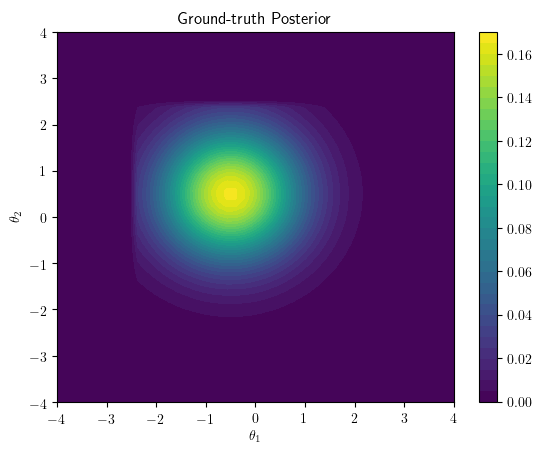}\\
      \includegraphics[width=0.48\textwidth]{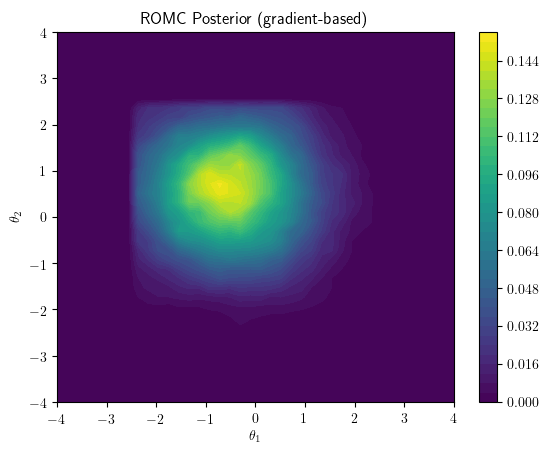}
      \includegraphics[width=0.48\textwidth]{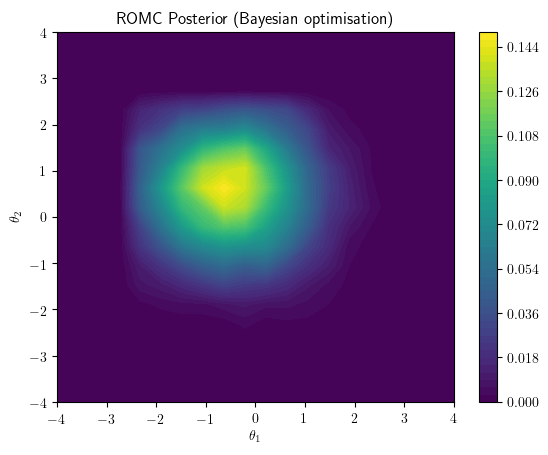}
    \end{center}
    \caption[2D example, evaluation of the approximate posterior.]{(a) First row: Ground-truth posterior approximated
      computationally. (b) Second row (left): ROMC approximate
      posteriors using gradient-based approach. The divergence from
      the ground-truth using the Jensen-Shannon distance is
      $0.068$. (c) Second row (right): ROMC approximate posterior
      using Bayesian optimisation. The divergence from the
      ground-truth using the Jensen-Shannon distance is $0.069$}
  \label{fig:ex2_4}
\end{figure}

Finally, the ground-truth and the approximate posteriors are presented
in figure \ref{fig:ex2_4}. We also confirm that the approximations are
close to the ground truth posteriors. As a note, we can observe that
the approximate posteriors present a diamond-shape in the mode of the
posterior; this happens due to the approximation of the circular
Gaussian-shape posterior with a sum of square boxes. The divergence
between the ground-truth distribution and the approximate ones is
$0.077$, using the Jensen-Shannon distance, which confirms the
satisfying matching between the two posteriors.

In this experiment we observed that the implementation fulfilled the
theoretical expectations for the ROMC inference method.

\begin{center} \label{tab:ex2}
\begin{tabular}{ c|c|c|c|c|c }
\hline
& $\mu_{\theta_1}$ & $\sigma_{\theta_1}$ & $\mu_{\theta_2}$ & $\sigma_{\theta_2}$ & Divergence\\
\hline \hline
Ground-truth & -0.45 & 0.935 & 0.45 & 0.935 & \\
\hline
ROMC (gradient-based) & -0.474 & 0.994 & 0.502 & 0.966 & 0.068\\
\hline
ROMC (Bayesian optimisation) & -0.485 & 0.987 & 0.511 & 0.939 & 0.069\\
\hline
\end{tabular}
\end{center}

In this simple artificial 2D example, with the ground-truth
information available, we confirmed that our implementation produces
accurate approximations. In the following section, we will question it
in a more involved case.

\subsection{Example 2: Second-order Moving Average MA(2)}
\label{subsec:ma2}
The second example is the second-order moving average (MA2) which is
used by the ELFI package as a fundamental model for testing all
inference implementations. This example is chosen to confirm that our
implementation of the ROMC approach produces sensible results in a
general model, since the previous ones where artificially created by
us.

\subsubsection*{Problem definition}

The second-order moving average (MA2) is a common model used for
univariate time series analysis. The observation at time $t$ is given by:

\begin{gather} \label{eq:ma2}
y_t = w_t + \theta_1 w_{t-1} + \theta_2 w_{t-2}\\
\theta_1, \theta_2 \in \R, \quad  w_k \sim \mathcal{N}(0,1), k \in \mathbb{Z}
\end{gather}

\begin{figure}[ht]
    \begin{center}
      \includegraphics[width=0.48\textwidth]{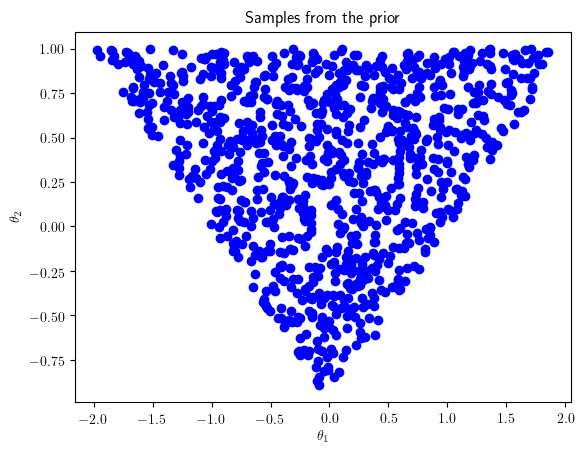}
    \end{center}
    \caption[MA2 example, prior distribution.]{Prior distribution proposed by \autocite{Marin2012}. The
      samples follow a triangular shape.}
  \label{fig:ma2_1}
\end{figure}

\noindent
The random variables $w_{k} \sim \mathcal{N}(0,1), k \in \mathbb{Z}$
represent an independent and identically distributed white noise and
$\theta_1, \theta_2$ the dependence from the previous
observations. The number of consecutive observations $T$ is a
hyper-parameter of the model; in our case we will set
$T=100$. Generating an MA2 time-series is pretty easy and efficient
using a simulator, therefore using a likelihood-free inference method
quite convenient. At the specific example, we use the prior proposed
by \autocite{Marin2012} for guaranteeing that the inference problem is
identifiable, i.e.\ loosely speaking the likelihood will have just one
mode. The multivariate prior, which is given in the equation
\eqref{eq:ma2_prior}, follows a triangular shape as plotted in figure
\ref{fig:ma2_1}.

\begin{equation} \label{eq:ma2_prior}
p(\thetab) = p(\theta_1)p(\theta_2|\theta_1)
= \mathcal{U}(\theta_1;-2,2)\mathcal{U}(\theta_2;\theta_1-1, \theta_1+1)
\end{equation}

\noindent
The vector $\yb_0 = (y_1, \ldots, y_{100})$ used as the observation,
has been generated with $\thetab=(0.6, 0.2)$. The dimensionality of
the output $\yb$ is quite large, therefore we use summary
statistics. Considering that the ouptput vector represents a
time-series signal, we prefer the autocovariance as the summary
statistic; we incorporate the autocovariances with $lag=1$ and
$lag=2$, as shown in equation \eqref{eq:ma2_summary}. Hence, the
distance is defined as the squared euclidean distance between the
summary statistics; this is the same choice as in \autocite{Marin2012}.

\begin{gather} \label{eq:ma2_summary}
  s_1(\yb) = \frac{1}{T-1} \sum_{t=2}^T y_ty_{t-1}\\
  s_2(\yb) = \frac{1}{T-2} \sum_{t=3}^T y_ty_{t-2} \\
  s(\yb) = (s_1(\yb), s_2(\yb))\\
  d = ||s(\yb) - s(\yb_0)||_2^2
\end{gather}

\subsubsection*{Perform the inference}

\begin{figure}[h]
    \begin{center}
      \includegraphics[width=0.48\textwidth]{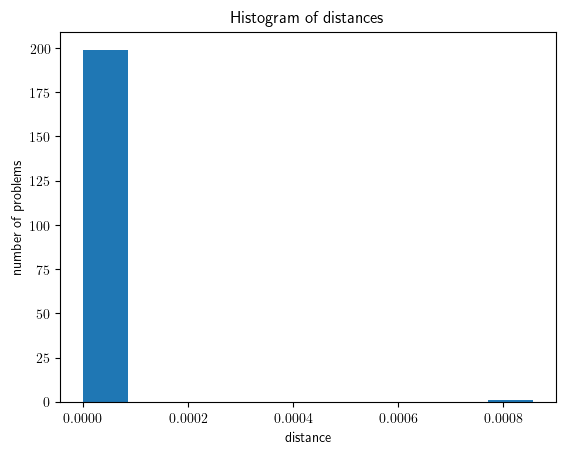}
      \includegraphics[width=0.48\textwidth]{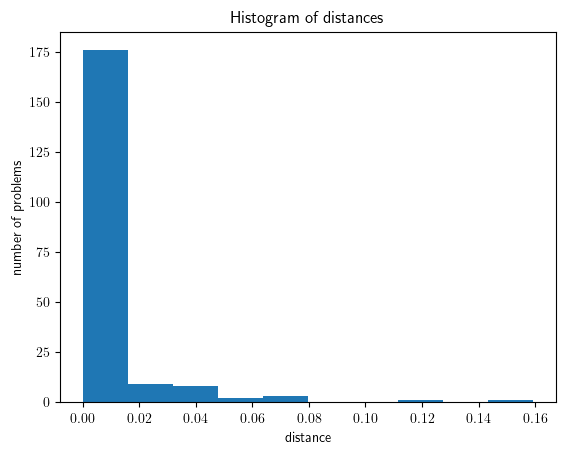}
    \end{center}
    \caption[MA2 example, histogram of distances]{Histogram of distances
      $d_i^*, i \in \{ 1, \ldots, n_1 \}$. The left graph corresponds
      to the gradient-based approach and the right one to the Bayesian
      optimisation approach.}
  \label{fig:ma2_2}
\end{figure}

As in the previous example, we perform the inference using the two
optimisation alternatives, the gradient-based and the Bayesian
optimisation. In this way, we compare the results obtained in each
step. For comparing the samples drawn with our implementation, we will
use the samples obtained with Rejection ABC. In figure \ref{fig:ma2_2}
we observe that in most cases the optimal distance $d_i^*$ is close to
zero in both optimisation approaches.

\begin{figure}[ht]
    \begin{center}
      \includegraphics[width=0.48\textwidth]{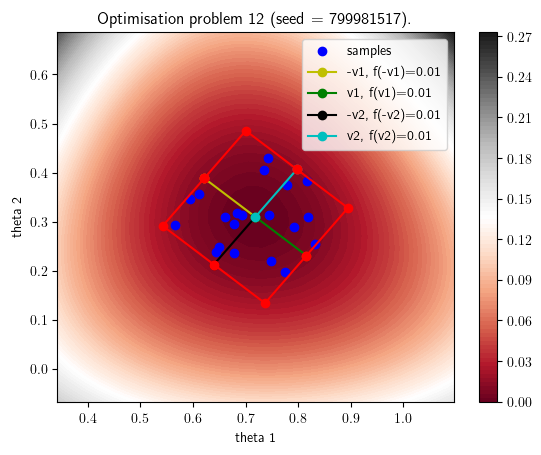}
      \includegraphics[width=0.48\textwidth]{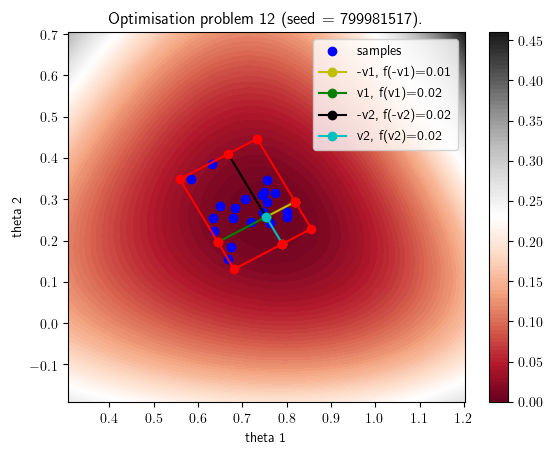}\\
      \includegraphics[width=0.48\textwidth]{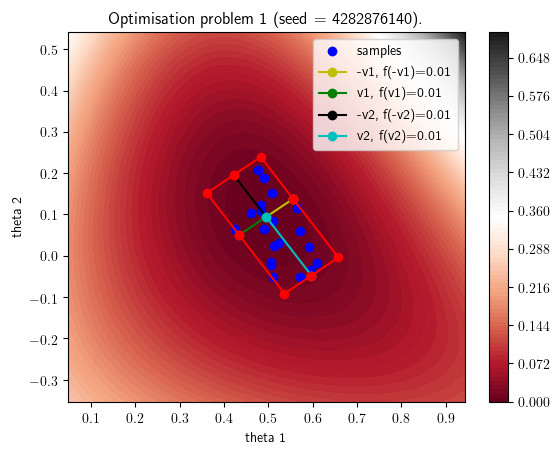}
      \includegraphics[width=0.48\textwidth]{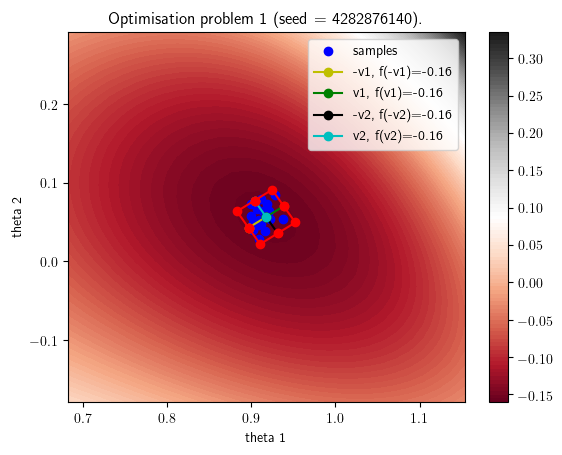}\\
      \includegraphics[width=0.48\textwidth]{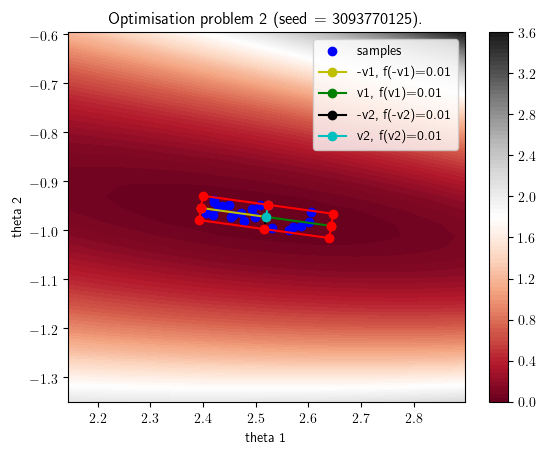}
      \includegraphics[width=0.48\textwidth]{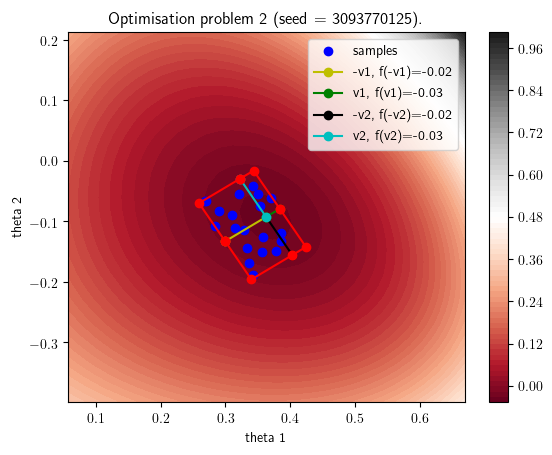}
    \end{center}
  \caption[MA2 example, the acceptance region in three distinct optimisation problems.]{Visualisation of the acceptance region in 3 different optimisation problems. Each row illustrates a different optimisation problem, the left column corresponds to the gradient-based approach and the right column to the Bayesian optimisation approach. The examples have been chosen to illustrate three different cases; in the first case, both optimisation schemes lead to similar optimal point and bounding box, in the second case the bounding box is similar in shape but a little bit shifted to the right relatively to the gradient-based approach and in the third case, both the optimal point and the bounding box is completely different.}
  \label{fig:ma2_5}
\end{figure}

\begin{figure}[ht]
    \begin{center}
      \includegraphics[width=0.48\textwidth]{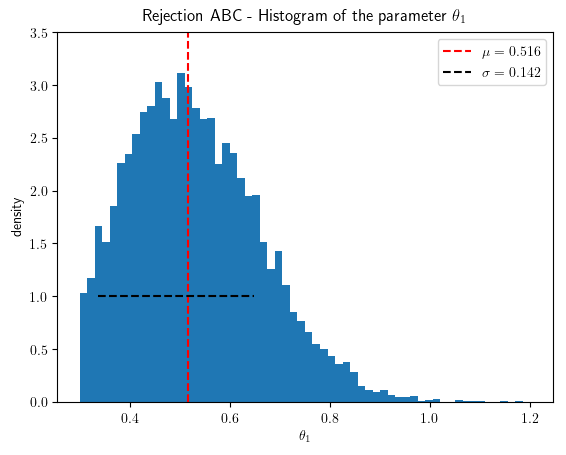}
      \includegraphics[width=0.48\textwidth]{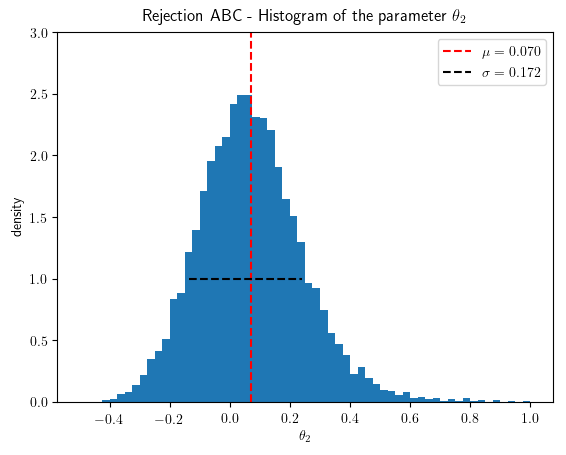}\\
      \includegraphics[width=0.48\textwidth]{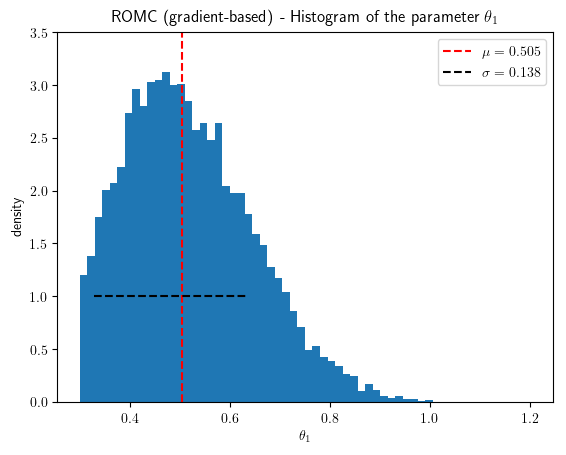}
      \includegraphics[width=0.48\textwidth]{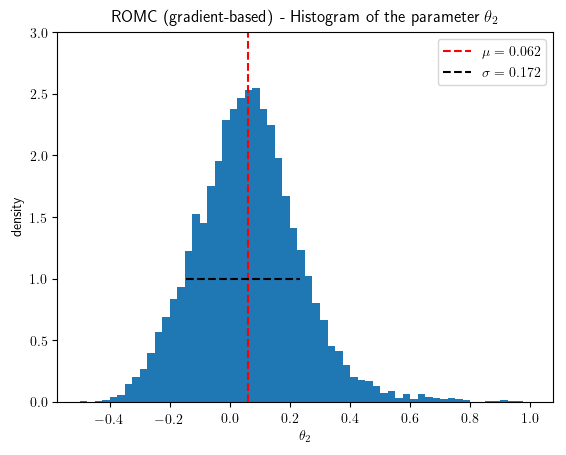}\\
      \includegraphics[width=0.48\textwidth]{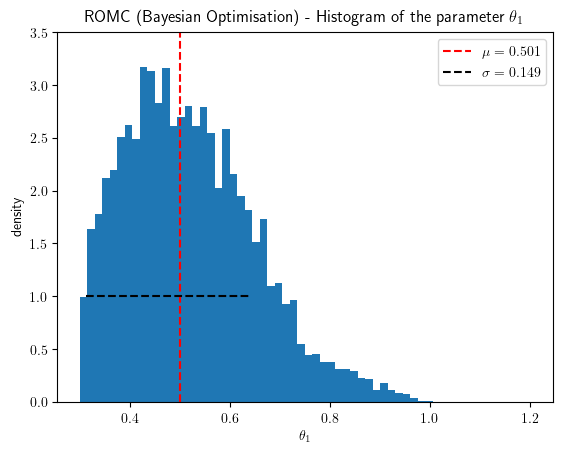}
      \includegraphics[width=0.48\textwidth]{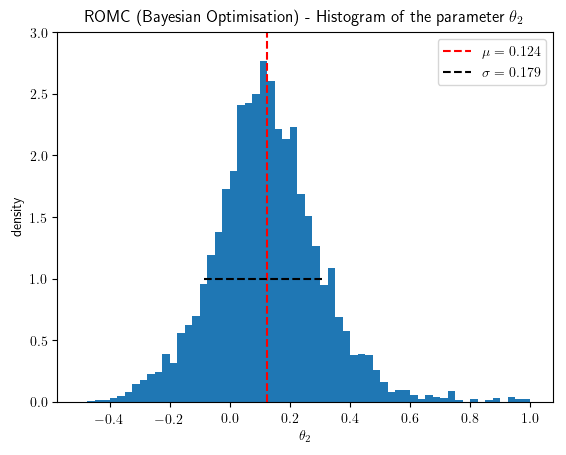}\\
    \end{center}
    \caption[MA2 example, evaluation of the marginal distributions.]{Histogram of the marginal distributions using
      different inference approaches; (a) in the first row, the
      approximate posterior samples are obtained using Rejection ABC
      sampling (b) in the second row, using ROMC sampling with
      gradient-based approach and (c) in the third row, using ROMC
      sampling with Bayesian optimisation approach. The vertical (red)
      line represents the samples mean $\mu$ and the horizontal
      (black) the standard deviation $\sigma$.}
  \label{fig:ma2_3}
\end{figure}

\begin{figure}[ht]
    \begin{center}
      \includegraphics[width=0.48\textwidth]{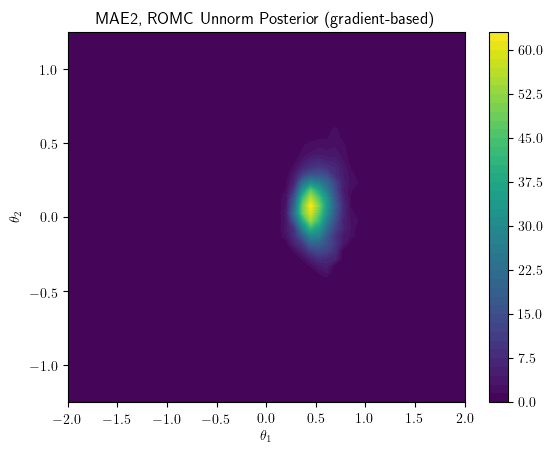}
      \includegraphics[width=0.48\textwidth]{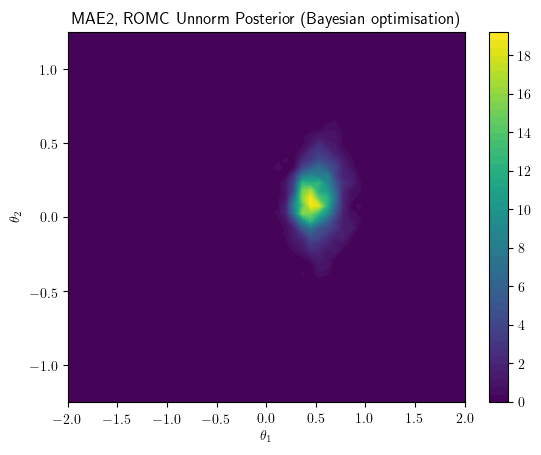}
    \end{center}
    \caption[MA2 example, approximate posterior.]{ROMC approximate posteriors using gradient-based approach
      (left) and Bayesian optimisation approach (right).}
  \label{fig:ma2_4}
\end{figure}

In figure \ref{fig:ma2_5}, we have chosen three different
deterministic optimisation cases (i.e. three different seeds) for
illustrating three different cases. In the first case, both
optimisation schemes lead to the creation of almost the same bounding
box. In the second case, the bounding box has a similar shape, though
different size and it is shifted along the $\theta_1$ axis. Finally,
in the third case, the bounding boxes are completely different. We can
thus conclude that although fitting a surrogate model has important
computational advantages, there is no guarantee that it will reproduce
accurately the local region around the optimal point. This
approximation error may lead to the construction of a considerably
different proposal region, which in turn, explains the differences in
the histogram of the marginal distributions presented in figure
\ref{fig:ma2_3} and in the approximate posteriors in figure
\ref{fig:ma2_4}.

In figure \ref{fig:ma2_3}, we demonstrate the histograms of the
marginal posteriors, using the three different methods; (a) Rejection
ABC (first line), (b) ROMC with gradient-based optimisation (second
line) and (c) ROMC with Bayesian optimisation (third
line). Undoubtedly, we observe that there is a significant similarity
between the three approaches. The Rejection ABC inference has been set
to infer 10000 accepted samples, with threshold $\epsilon=0.1$. The
large number of samples and the small distance from the observations
let us treat them as ground-truth information. In the table \ref{tab:ma2}
we present the empirical mean $\mu$ and standard deviation $\sigma$
for each inference approach. We observe that there is a significant
agreement between the approaches, which verifies that the ROMC
implementation provides sensible results. Finally, in figure
\ref{fig:ma2_4} we provide the ROMC approximate posteriors using
gradients and Bayesian optimisation; as confirmed by the statistics in
\ref{tab:ma2}, both posteriors have a single mode, located at the same
point, with a larger variance observed in the Bayesian Optimisation
case. The posteriors mode is quite close to the parameter
configuration that generated the data i.e.\ $\thetab = (0.6, 0.2)$.

\begin{center} \label{tab:ma2}
\begin{tabular}{ c|c|c|c|c }
\hline
& $\mu_{\theta_1}$ & $\sigma_{\theta_1}$ & $\mu_{\theta_2}$ & $\sigma_{\theta_2}$ \\
\hline \hline
Rejection ABC & 0.516 & 0.142 & 0.07 & 0.172 \\
\hline
ROMC (gradient-based) & 0.495 & 0.136 & 0.048 & 0.178 \\
\hline
ROMC (Bayesian optimisation) & 0.510 & 0.156 & 0.108 & 0.183 \\
\hline
\end{tabular}
\end{center}

\subsection{Execution Time Experiments}
\label{subsec:exec}
In this section, we will present the execution time of the basic ROMC
functionalities. Apart from performing the inference accurately, one
of the notable advantages of ROMC is its efficiency. In terms of
performance, ROMC holds two key advantages.

Firstly, all its subparts are parallelisable; optimising the objective
functions, constructing the bounding boxes, fitting the local
surrogate models, sampling and evaluating the posterior can be
executed in a fully-parallel fashion. Therefore, the speed-up can be
extended as much as the computational resources allow. Specifically,
at the CPU level, the parallelisation can be incorporate all available
cores. A similar design can be utilised in the case of having a
cluster of PCs available. Parallelising the process at the GPU level
is not that trivial, since the parallel tasks are more complicated
than simple floating-point operations. Our current implementation
supports parallelisation only at the CPU level, exploiting all the cores of the CPU.\footnote{At a future update, it is possible to support parallelising the processes using a cluster of PCs.}. In subsection
\ref{subsubsec:parallel}, we will demonstrate the speed-up achieved
through parallelisation.

\begin{figure}[ht]
    \begin{center}
      \includegraphics[width=0.48\textwidth]{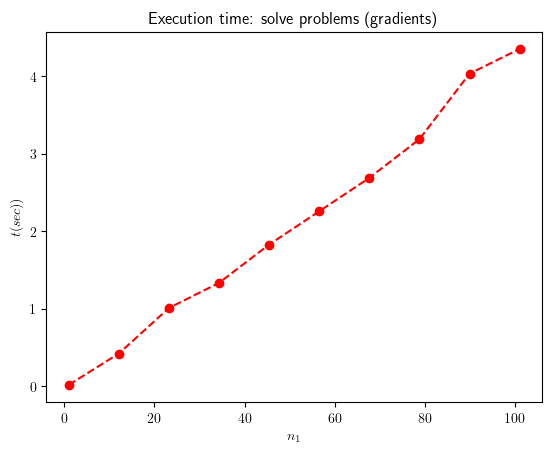}
      \includegraphics[width=0.48\textwidth]{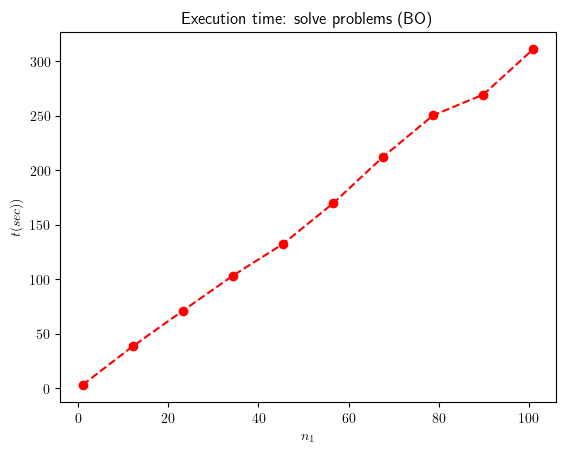}
    \end{center}
    \caption[Execution time for solving the optimisation problems.]{Execution time for defining and solving the optimisation
      problems. We observe that the Bayesian optimisation scheme is much more expensive, performing the task slower by a factor of $75$.}
  \label{fig:exec_solve}
\end{figure}

\begin{figure}[ht]
    \begin{center}
      \includegraphics[width=0.48\textwidth]{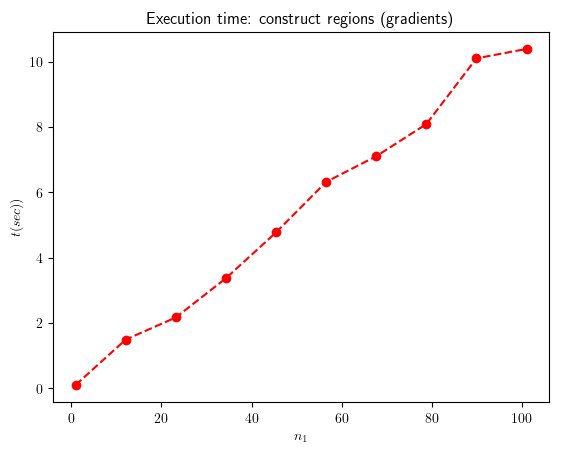}
      \includegraphics[width=0.48\textwidth]{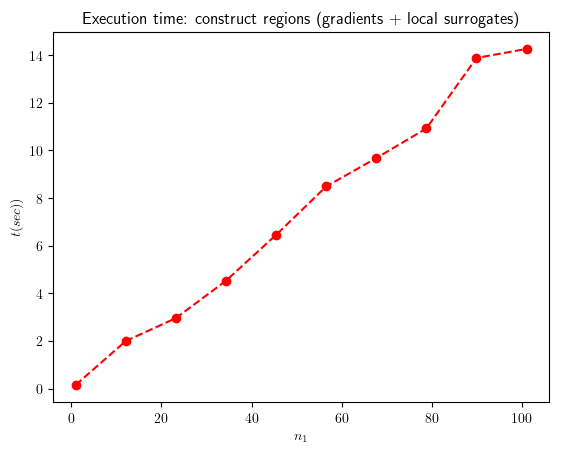}\\
      \includegraphics[width=0.48\textwidth]{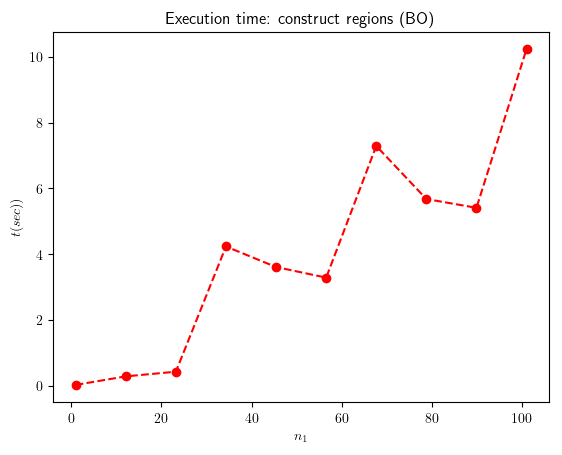}
      \includegraphics[width=0.48\textwidth]{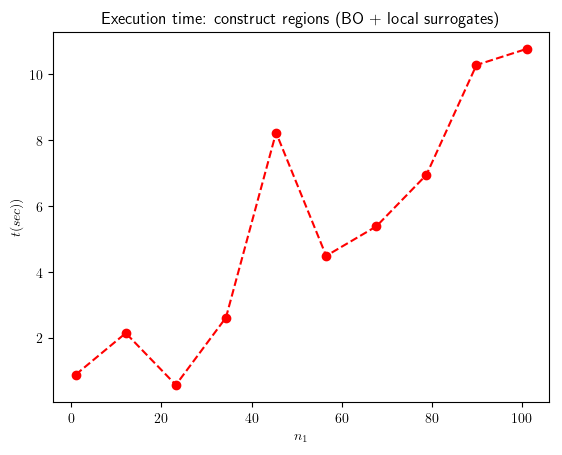}
    \end{center}
    \caption[Execution time for constructing the n-dimensional bounding box regions.]{Execution time for constructing the n-dimensional
      bounding box region and, optionally, fitting the local surrogate
      models. We observe that fitting the surrogate models incurs a
      small increase by a factor of $1.5$.}
  \label{fig:exec_regions}
\end{figure}

\begin{figure}[ht]
    \begin{center}
      \includegraphics[width=0.48\textwidth]{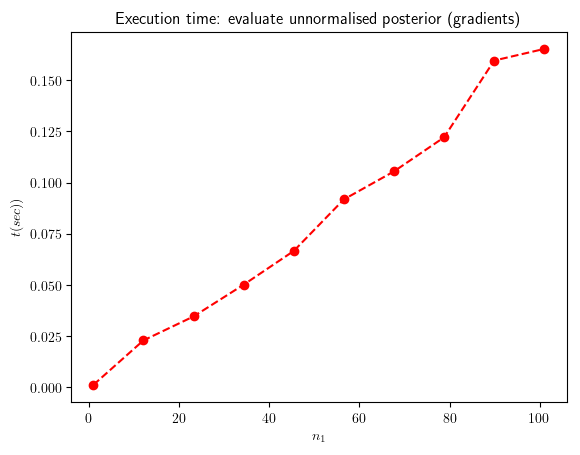}
      \includegraphics[width=0.48\textwidth]{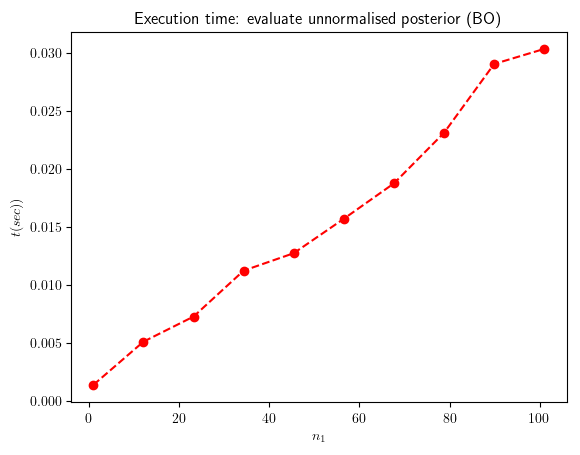}\\
      \includegraphics[width=0.48\textwidth]{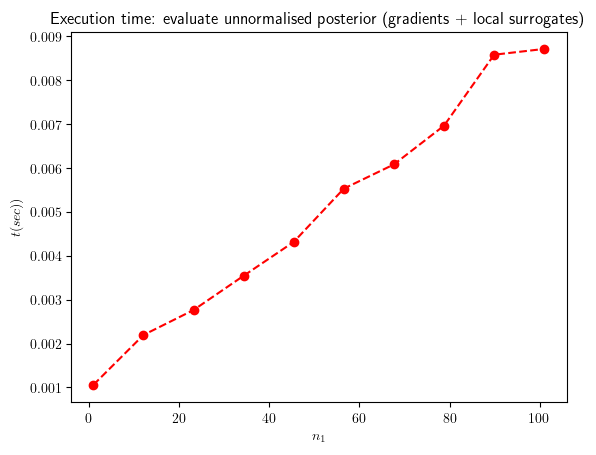}
    \end{center}
    \caption[Execution time for evaluation the unnormalised posterior.]{Execution time for evaluating the unnormalised posterior
      approximation. We observe that using a Gaussian Process surrogate model is almost 5 times faster than the simulator and the quadratic local surrogate model 15 times faster.}
  \label{fig:exec_posterior}
\end{figure}

\begin{figure}[ht]
    \begin{center}
      \includegraphics[width=0.48\textwidth]{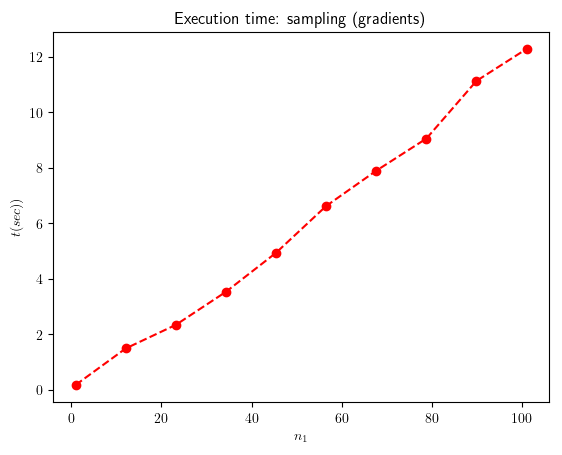}
      \includegraphics[width=0.48\textwidth]{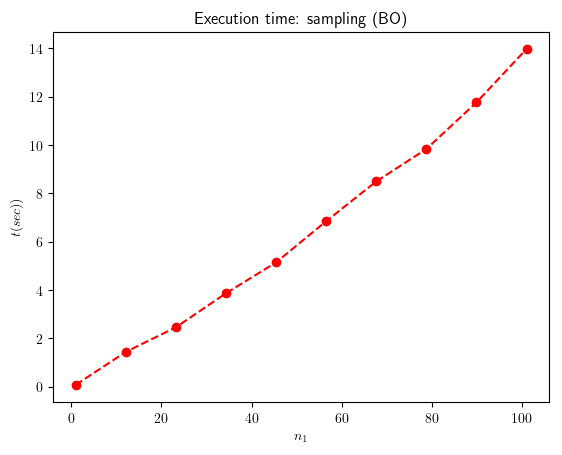}\\
      \includegraphics[width=0.48\textwidth]{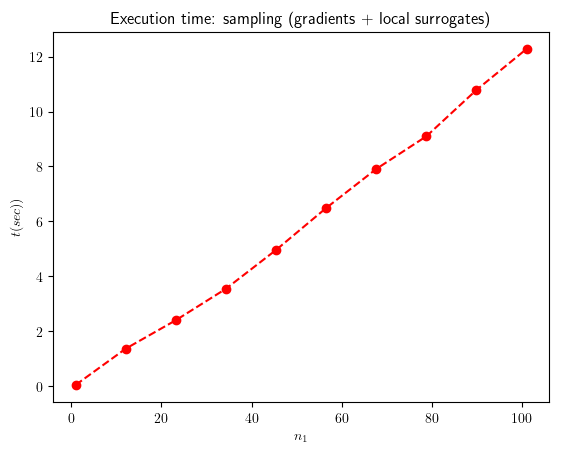}
    \end{center}
    \caption[Execution time for sampling from the approximate posterior.]{Execution time for sampling from the approximate
      posterior. We observe a small speed-up using the local surrogate model.}
  \label{fig:exec_sample}
\end{figure}

The second important advantage concerns the execution of the
training and the inference phase in distinct time-slots. Therefore, one
can consume a lot of training time and resources but ask for
accelerated inference. The use of a lightweight surrogate model around
the optimal point exploits this essential characteristic; it trades
some additional computational burden at the training phase, for
exercising faster inference later. In figures \ref{fig:exec_solve},
\ref{fig:exec_regions}, \ref{fig:exec_posterior},
\ref{fig:exec_sample} we can observe this advantage. The
example measured in these figures is the simple 1D example, used in
the previous chapter. We observe that fitting local surrogate models
slows down the training phase (estimating the regions) by a factor of
$1.5$, but provides a speed-up of $15$ at the inference phase i.e.\
approximating unnormalised posterior. This outcome would be even more
potent in larger models, where running the simulator is even more
expensive.

\subsubsection{The effect of parallelisation} \label{subsubsec:parallel}

\begin{figure}[ht]
    \begin{center}
      \includegraphics[width=0.48\textwidth]{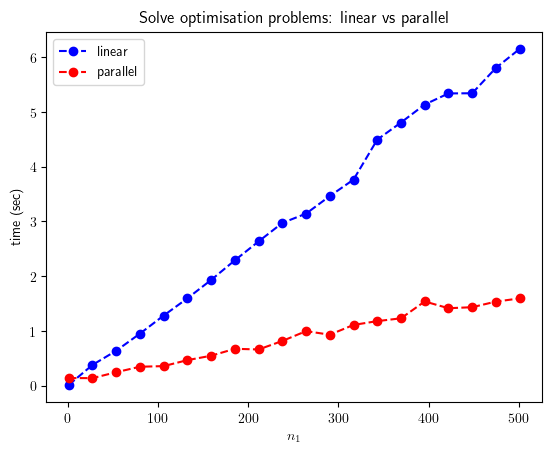}
      \includegraphics[width=0.48\textwidth]{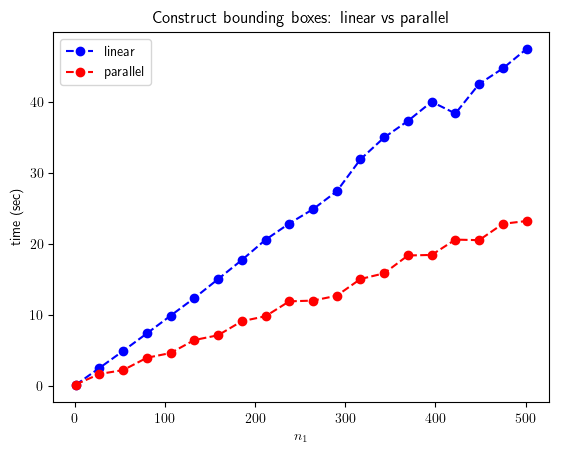}
    \end{center}
    \caption[Execution time of the training part exploiting parallelisation]{Execution
      time of the training part exploiting parallelisation. At the left figure, we measure
      the execution time for solving the optimisation problems. At the
      right figure, we measure the execution time for constructing the
      bounding boxes.}
  \label{fig:exec_parallel}
\end{figure}

\begin{figure}[ht]
    \begin{center}
      \includegraphics[width=0.48\textwidth]{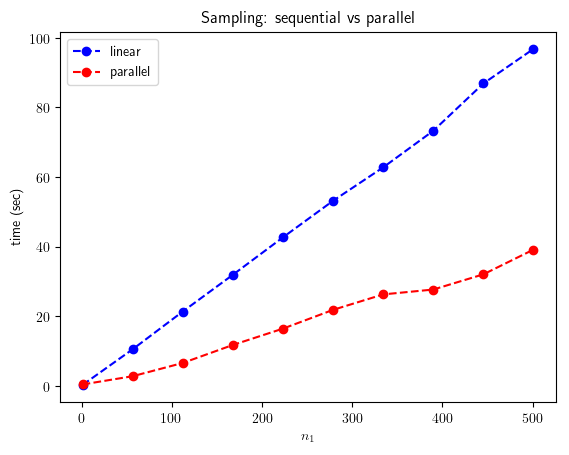}
      \includegraphics[width=0.48\textwidth]{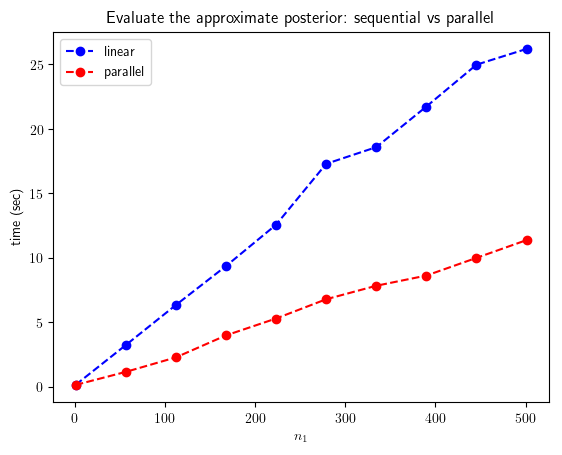}
    \end{center}
    \caption[Execution time of the inference part exploiting parallelisation.]{Execution
      time of the inference part exploiting parallelisation. At the left figure, we measure
      the execution time for sampling $n_2=50$ points per region. At the right figure we measure the execution time for evaluating the posterior at a batch of $50$ points.}
  \label{fig:exec_parallel2}
\end{figure}

In this section, we measure the execution times for (a) solving the optimisation problems, (b) construct the bounding boxes, (c) sample from the posterior and (d) evaluate the approximate posterior using parallelisation. The model used for this experiment is the simple two-dimensional example of section \ref{subsec:ex1}. All experiments
have been executed in a laptop Dell XPS 15 with an 8-core CPU. Therefore the maximum expected speed-up can reach a factor of 8. Normally the observed speed-up is lower due to the overhead of setting-up the parallel processes. The experiments confirm our general expectation; the parallel version performs all tasks between 2.5 and 6 times faster compared to the sequential, as shown in figures \ref{fig:exec_parallel} and \ref{fig:exec_parallel2}. The most benefited task is solving the optimisation problems which is performed almost 6 times faster. Sampling is executed 3.5 times faster, whereas evaluating the posterior and constructing the bounding boxes almost 2.5 times faster. 

\clearpage

\section{Conclusion}

This chapter concludes the dissertation. We summarise the main
contributions of our work, both on the theoretical and the
implementation side. Furthermore, based on our observation throughout
the implementation of the ROMC method, we suggest some ideas for
future research.

\subsection{Outcomes}
\label{subsec:outcomes}
In this dissertation, we have studied the likelihood-free inference
approaches, focusing on the novel ROMC method that we implemented in
an open-source software package. The main contribution of the
dissertation is this implementation which can be used mainly by the
research community for further experimentation.

In chapter 2, we presented the simulator-based models explaining the
particularities of the inference when the likelihood is not
tractable. We then presented an overview of the Optimisation
Monte-Carlo methods (OMC and ROMC) examining their strategies on
approximating the posterior. We discussed the point-of-view of the
ROMC approach, demonstrating the mathematic modelling it
introduces. We also presented the aspect of ROMC as a
meta-algorithm. Finally, at the end of chapter 2, we transformed the
mathematical modelling of ROMC into algorithmic form. Up to this
point, the dissertation mainly restates the ideas presented in the
original paper \autocite{Ikonomov2019} and, hence, it can be used
together with the paper by a reader who wants to understand the ROMC
approach.

The notable contribution of the dissertation is the implementation of
the method in the ELFI package. Due to the novelty of the method, it
has not been implemented in any other package so far. We tried to
implement the method focusing on four principles: simplicity,
accuracy, efficiency and extensibility. We tried to provide the user
with a simple-to-use method. Therefore, we followed the guidelines of
the ELFI package, aligning ROMC with the other provided inference methods. We
also kept the function calls simple, asking only for the necessary
arguments. Secondly, we tested the implementation on a range of
examples for ensuring accurate inference results. We designed some
artificial examples for evaluating our implementation using
ground-truth information. We also tested that it works smoothly under
general models, not artificially created by us. Thirdly, we tried to
solve the tasks efficiently. We applied parallel processing which offers a major speedup in the the execution time of all tasks. Furthermore, we avoided using redundant calls, we
exploited vectorisation when possible, and we avoided expensive
for-loops.  We measured the execution time of the method for providing
an overview of the time needed for performing each task. Finally, we
preserved extensibility; this was a significant priority in the
implementation design. Apart from offering a ready-to-use inference
method, we want our implementation to serve as the initial point for
researchers who would like to further experiment with the method. This
requirement aligns with the nature of ROMC as a meta-algorithm; one
can replace the method involved in a specific task without the rest of
the algorithm to collapse. Finally, we provide extensive documentation
of all implemented methods and a collection of examples for
illustrating the main use-cases. We wanted the reader to be able to
observe and interact in-practice with the functionalities we offer.

\subsection{Future Research Directions}
\label{subsec:future}
Future research directions involve both the theoretical and
implementation aspect of the ROMC approach.

In the theoretical part, an essential drawback of the ROMC approach is
the creation of a single bounding box for each objective
function. This simplification can be error-prone in situations where
we have ignorance about the properties of the objective
function. Consider, for example, the scenario where an objective
function defines multiple tiny disjoint acceptance areas; the
optimiser will (hopefully) reach only one, and a bounding box will be
constructed around it, compressing all the mass of the distribution
there. Therefore, this small region will erroneously dominate the
specific objective function. In general, the existence of multiple
disjoint acceptance regions complicates the inference. A naive
approach for improving this failure would be solving each optimisation
problem more than once, using different starting points. With this
approach, we hope that the optimiser will reach a different local
minimum in each execution. However, apart from the added computational
complexity, this approach does not guarantee an improvement. In
general, identifying all the acceptance regions in an unconstrained
problem with many local optimums is quite challenging.

Further research on the optimisation part, may also focus on modelling
the case where the parameter space is constrained. Consider, for
example, the case where a parameter represents a probability (as in
the tuberculosis example). In this case, the parameter is constrained
in the region $[0,1]$. Defining a prior with zero mass outside of this
area ensures that all non-acceptable values will not contribute to the
posterior. However, the optimiser does not consider the prior. Hence,
we can observe the phenomenon where the majority of the (or all)
acceptance regions is cancelled-out by the prior. In this case, the
posterior will be dominated by a few acceptance region parts that
overlap with the prior, which is not robust. Dealing with such
constrained non-linear optimisation problems, which is particularly
challenging, can serve as a second research line.

The previous two research directions overlap with the field of
optimisation. Another research avenue, more concentrated on the ROMC
method, is the approximation of the acceptance regions with a more
complex geometrical shape than the bounding box. In the current
implementation, the bounding box only serves as the proposal region
for sampling; running the indicator function afterwards, we manage to
discard the samples that are above the distance threshold. This
operation is quite expensive from a computational point of view. The
workaround proposed from the current implementation is fitting a local
surrogate model for avoiding running the expensive simulator. However,
fitting a surrogate model for every objective function is still
expensive. Approximating the acceptance region accurately with a more
complicated geometrical structure could remove the burden of running
the indicator function later. For example, a simple idea would be
approximating the acceptance region with a set of non-overlapping
bounding boxes. We insist on using the square as the basic shape
because we can efficiently sample from a uniform distribution defined
on it. A future research direction could investigate this capability.

On the implementation side, the improvements are more concrete. A critical
improvement would be performing the training and the inference phase
in a batched fashion. For example, the practitioner may indicate that
he would like to approximate the posterior in batches of N
optimisation problems; this means that the approximation should be
updated after each N new bounding boxes are constructed. Utilising
this feature, the user may interactively decide when he would like to
terminate the process. Finally, the implementation could also support
saving the state of the training or the inference process for
continuing later. In the current implementation, although the user can
save the inference result, he cannot interrupt the training phase and
continue later.
\clearpage

\printbibliography
\clearpage

% \appendix
% \section*{Appendices}
% \addcontentsline{toc}{section}{Appendices}

% \clearpage
% \section{An Appendix}
% \label{app:one}

% \clearpage

% \section{Another Appendix}
% \label{app:two}

\end{document}